\documentclass[11pt,draftcls,onecolumn]{IEEEtran}
\usepackage{booktabs}
\usepackage{setspace}
\usepackage[numbers, square, comma, sort&compress]{natbib}
\usepackage{bm}
\usepackage{shortcuts}
\usepackage{subfigure}
\usepackage{amsthm, amsmath, amssymb, dsfont, epsfig, pifont, algorithmic, algorithm, psfrag, color}
\usepackage{graphicx}
\newcommand{\ra}[1]{\renewcommand{\arraystretch}{#1}}
\newcommand{\myqed}{\hfill \ensuremath{\blacksquare}}

\title{Calibration Using Matrix Completion with Application to Ultrasound Tomography}
\author{Reza Parhizkar$^*$, \IEEEmembership{Student Member, IEEE}, Amin Karbasi, \IEEEmembership{Student Member, IEEE},  \\ Sewoong Oh, \IEEEmembership{Student Member, IEEE}, and Martin Vetterli, \IEEEmembership{Fellow, IEEE}

\thanks{ Reza Parhizkar, Amin Karbasi and Martin Vetterli are with the School of Computer and Communication
Sciences, Ecole Polytechnique F\'{e}d\'{e}rale de Lausanne (EPFL), CH-1015
Lausanne, Switzerland (e-mails: \{reza.parhizkar, amin.karbasi, martin.vetterli\}@epfl.ch). Sewoong Oh is with the Laboratory for Information and Decision Systems, MIT, MA 02139, USA (e-mail: swoh@mit.edu). Martin Vetterli is also with the Department of Electrical Engineering and Computer Sciences, University of California, Berkeley, CA 94720, USA.}%
\thanks{This work has been partly presented in the conference proceeding \cite{par11}.}
\thanks{This work was supported by the Swiss National Science Foundation under grants 200021-121935 and ERC Advanced Grant--Support for Frontier Research--SPARSAM, Nr: 247006.}
}

\begin{document}
%\onehalfspacing
  \maketitle
%  \IEEEkeywords{ultrasound tomography, calibration, matrix completion, multi-dimensional scaling}
  \begin{abstract}
We study the calibration process in circular ultrasound tomography devices where the sensor positions deviate from the circumference of a perfect circle. This problem arises in a variety of  applications in signal processing ranging from breast imaging  to sensor network localization. We introduce a novel method of calibration/localization based on the time-of-flight (ToF) measurements between sensors when the enclosed medium is homogeneous. In the presence of all the pairwise ToFs, one can easily estimate the sensor positions using multi-dimensional scaling (MDS) method.  In practice however, due to the transitional behaviour of the sensors and the beam form of the transducers, the ToF measurements for close-by sensors are unavailable. Further, random malfunctioning of the sensors leads to random missing ToF measurements. On top of the missing entries, in practice an unknown time delay is also added to the measurements. In this work, we incorporate the fact that a matrix defined from all the ToF measurements is of rank at most four. In order to estimate the missing ToFs, we apply a state-of-the-art low-rank matrix completion algorithm, \opt.  To find the correct positions of the sensors (our ultimate goal) we then apply MDS. We show analytic bounds on the overall error of the whole process in the presence of noise and hence deduce its robustness. Finally, we confirm the functionality of our method in practice by simulations mimicking the measurements of a circular ultrasound tomography device. 

\end{abstract}
\begin{center}
\textbf{EDICS Category: SAM-CALB, BIO-SENS, SAM-IMGA, SEN-LOCL}
\end{center}
%\begin{IEEEkeywords}
%ultrasound tomography, calibration, sensor localization, matrix completion, multi-dimensional scaling, breast imaging
%\end{IEEEkeywords}
%%%%%%%%%%%%%%%%%%%%% Introduction %%%%%%%%%%%%%%%%%%
\begin{section}{Introduction}
\label{sec:introduction}

In most applications involving sensing, finding the correct positions of the sensors is of crucial importance for obtaining reliable results. This is particularly true in the case of inverse problems which can be very sensitive to incorrect sensor placement. This requirement can be satisfied in two ways; One might put the effort in the construction of the instruments and try to place the sensors exactly in the desired positions, or use a method to find the exact positions after the construction of the device. In this work we will consider the latter and we call the procedure of obtaining the sensor positions calibration. Note that even in the former case, due to  the precision of the construction instruments, a calibration is needed afterwards for determining the exact sensor positions. Although in rare cases a single calibration might be enough throughout the lifetime of the measurement system, it is of great use to have a calibration procedure which can be repeated easily and with low cost.

This work focusses on the calibration problem in circular sensing devices, in particular, the ones manufactured and deployed in ~\cite{dur07, jov09}.
 %the assumed model is based on the circular tomography devices which are deployed in ~\cite{dur07, jov09}. 
 These devices consist of a circular ring surrounding an object and scanning horizontal planes. Ultrasound sensors are placed on the interior boundary of the ring and act both as transmitters and receivers.
 
The calibration problem we address in this paper is the following; In the circular tomography devices, the sensors are not exactly placed on a perfect circle. This uncertainty in the positions of the sensors acts as a source of error in the reconstruction algorithms used to obtain the characteristics of the enclosed object. We aim at finding a simple method for calibrating the system with correct sensor positions with low cost and without using any extra calibrating instrument.

In order to find the correct sensor positions, we incorporate the time-of-flight (ToF) of ultrasound signals between pairs of sensors, 
which is the time taken by an ultrasound wavefront to travel from a transmitter to a receiver. 
If we have all the ToF measurements between all pairs of sensors 
when the enclosed medium is homogeneous, then we can construct a ToF matrix 
where each entry corresponds to the ToF between each pair of sensors. 
We can infer the positions of the sensors using this ToF matrix.

To obtain reliable ToF entries appropriate for our purpose, we assume that no object is placed inside the ring during the calibration phase and prior to actual measurements.  There are a number of challenges we are encountering in this work, namely, 
\begin{itemize}
\item the ToF matrices obtained in a practical setup have missing entries. 
\item the measured entries of the ToF matrices are corrupted by noise.
\item there is an unknown time delay added to the measurements.
\end{itemize}

If one had the complete and noiseless ToF matrix without time delay, the task of finding the exact positions would be very simple. This problem is addressed in  literature as multi-dimensional scaling (MDS) \cite{dri06}. Unfortunately, the ToF matrix in practical setups is never complete and many of the time-of-flight values are missing. The missing entries can be  divided into two categories; \textit{structured missing entries} caused by inability of the sensors to compute the mutual time-of-flights with their close-by neighbors,  and \textit{random missing entries} due to malfunctioning of the sensors or the ToF estimation algorithm during the measurement procedure. 

A good estimation of the positions of the sensors 
can be obtained, if we have a good estimation of 
the missing entries of the ToF matrix. 
In general, it is a difficult task to infer missing entries of a matrix. 
However, it has recently been established that if the matrix is low rank, 
a small random subset of its entries permits an exact reconstruction \cite{candes08}.
%This result was first proved by Cand\`{e}s and Recht who analysed a convex relaxation 
%of this low-rank matrix completion problem \cite{candes08}. 
%More recently, an alternative approach using a combination of 
%spectral techniques and manifold optimization was introduced in \cite{KMO09exact}. 
%This novel algorithm used in our work is referred as {\sc OptSpace} and 
%has been shown to be stable under noisy measurements \cite{KMO09noise}. 
Since a modified version of the ToF matrix (when the entries are squared ToF measurements) is low rank, its missing entries can be accurately estimated using matrix completion algorithms. To this end,  we use \opt,  a robust matrix completion algorithm developed by Keshavan et al. \cite{KMO09noise}.

%The main challenge we are facing in this work is to acquire a good guess about missing entries. In general, the missing entries of a matrix can not be accurately estimated. However, it has recently been discovered that if the matrix is low rank,  a small random subset of its entries allow to reconstruct it exactly. This result was first proved by Candes and Recht by introducing a convex relaxation \cite{candes08} and alternatively solved by authors of 
%\cite{KMO09exact} using a combination of spectral techniques and manifold optimization. The algorithm proposed in \cite{KMO09exact} and used in our work is referred as  {\sc OptSpace} and has been shown to be stable under noisy measurements\cite{KMO09noise}. 

%Motivated by this result, we show that a matrix formed by ToF values is low rank and hence by using {\sc OptSpace} its missing entries can be accurately estimated. 
 
On top of the missing entries, we also need to deal with an unknown time delay. This delay it due to the fact that in practice, the impulse response of the piezoelectric and the time origin in the measurement procedure are not known, and this causes an unknown time delay which should then be added to the measurements. To infer this time delay simultaneously with the positions of the sensors, 
we propose a heuristic algorithm based on {\sc OptSpace}.

In circular setups, the sensors are not necessarily on a circle and deviate from the circumference which in fact motivates the calibration problem. We therefore need to assume that they are in the proximity of a circle (the precise statement is given later) and we are required to find the exact positions. Our approach is to estimate the local and random missing pairwise distances from which we can then infer the positions.
%In this work, we assume that the sensors are in the proximity of a circle and local distance measurements are missing. 
As we have already mentioned, we show that a modified version of the ToF matrix has  rank at most four, and using this property, we propose our calibration procedure. The block diagram shown in Fig.~\ref{fig:calib_block_diag} summarizes the procedure.

\begin{figure}[tb]
\centering
\psfrag{denoising}{denoising}
\includegraphics[width = \linewidth]{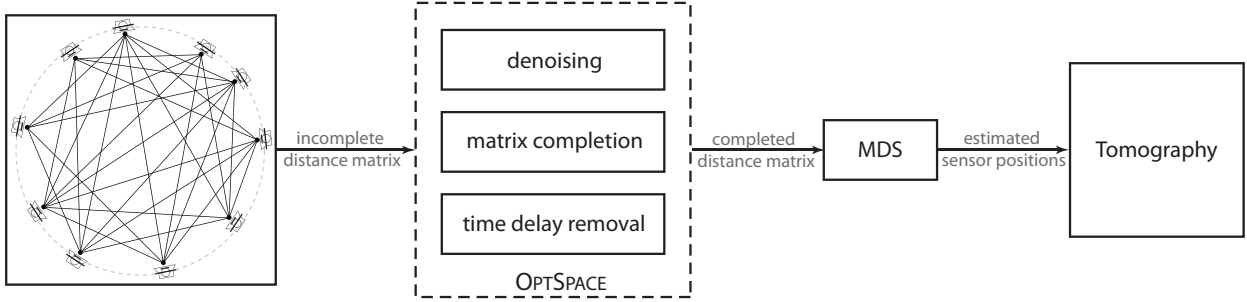}
\caption{Block diagram for the calibration procedure prior to ultrasound tomography. The incomplete distance matrix is passed through the \opt algorithm which denoises it, estimates the missing entries and removed the unknown time delay. The calibration is finished then by applying the MDS algorithm on the completed matrix which estimates the actual sensor positions. }
\label{fig:calib_block_diag}
\end{figure}

\subsection{Related work}
\label{sec:relatedWork}
Calibration for circular tomography devises is a variant of sensor localization, a problem  that has been extensively studied for the past decade \cite{range-free, GPS-free}. In sensor localization, given the\textit{ local connectivity} (i.e., which sensors are in the communication range of which others), the objective is to devise an algorithm that can infer the global position of the sensors. In practice, several methods are deployed as a means of obtaining this local information:  the Signal Strength \cite{signal-strength}, the Angle of Arrival (AOA) \cite{AOA}, and the Time Difference of Arrival (TDOA) \cite{TDOA}. Our problem is naturally related to sensor localization when estimated TDOAs are used to measure the pairwise distances between nearby nodes. One should note that due to energy constraints, each node has a small communication range compared to the field size  they are installed. As a result,   only nodes within the communication range of each other can communicate and hence estimate their pairwise TDOA's.  This situation is depicted in Fig.~\ref{localization-calibration}.

In our problem, however, the local connectivity is precisely the kind of information that is missing. In fact, the beam width of transducers and the transition of ultrasound sensors  disallow us from having reliable ToF's for nearby sensors (see Section \ref{sec:tof_estim}). For this reason, in practice, the ToF's for close-by sensors are discarded and no information regarding their pairwise distances can be deduced. Consequently, in our scenario we are faced with  a  different setting from that of sensor localization; namely, 
\begin{itemize}
\item the pairwise distances of neighboring sensors are missing,
\item only the pairwise distances of faraway sensors can be figured out from their ToF's.
\end{itemize}
This situation is demonstrated  in Fig.~\ref{localization-calibration}. By comparing these two scenarios in Fig.~\ref{localization-calibration}, one can think of the calibration problem for ultrasound sensors as the dual problem of sensor localization.  As a result, all sensor localization algorithms that rely on local information/connectivity are doomed to fail in our scenario. To confirm this fact, in Section \ref{sec:experimental_results} through numerical simulations we compare the performance of our proposed method with the state-of-the-art algorithms for sensor localization applied in our setting. 

\begin{figure}[t!]
\centering
\subfigure[]{
\psfrag{denoising}{denoising}
\psfrag{a}[l][l]{unknown}
\psfrag{b}[l][l]{distance}
\psfrag{c}[l][l]{known}
\psfrag{d}[l][l]{\footnotesize radio range}
\includegraphics[scale=0.2]{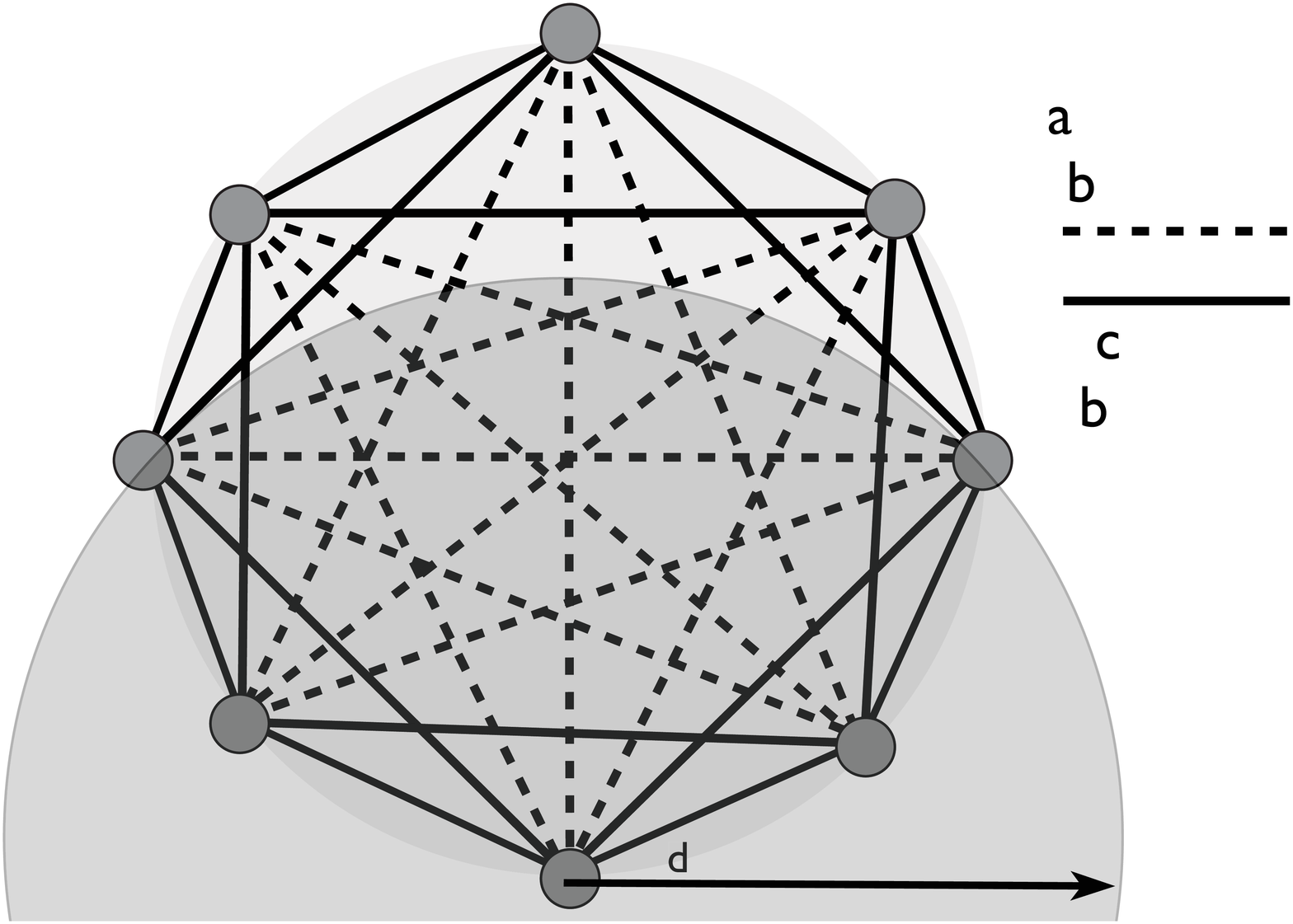}
\label{subfig:radio}}
\subfigure[]{\psfrag{denoising}{denoising}
\psfrag{a}[l][l]{unknown}
\psfrag{b}[l][l]{distance}
\psfrag{c}[l][l]{known}
\psfrag{d}[l][l]{\footnotesize radio range}
\includegraphics[scale=0.2]{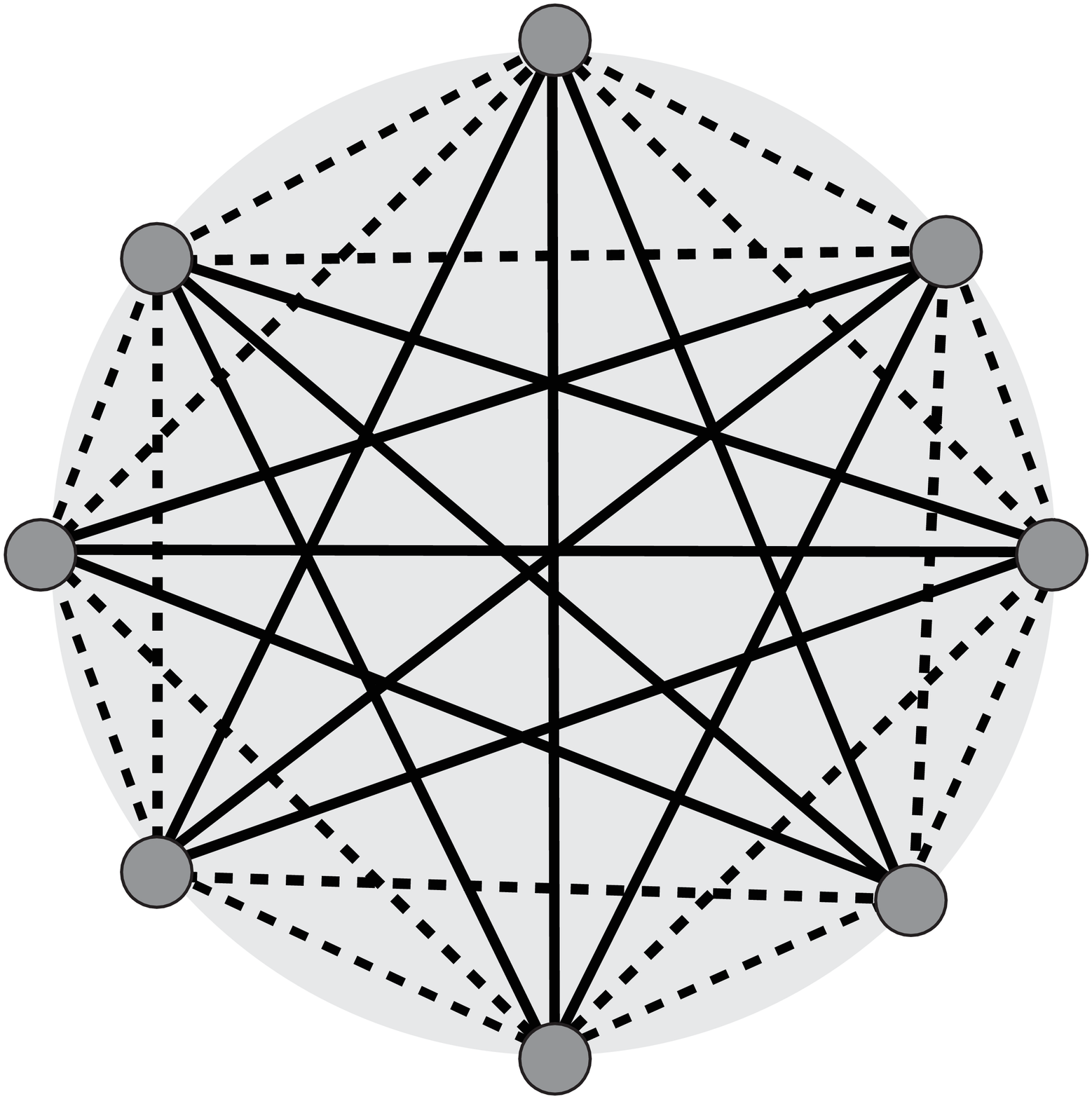}
\label{subfig:calib}}
\caption{In sensor localization (left figure) the local connectivity information is available and faraway ones are missing whereas in calibration (right figure) the opposite is true.}
\label{localization-calibration}
\end{figure}

The first sensor localization algorithm we consider is \textsc{Mds-Map} \cite{SRZ03}. This algorithm has two phases. First, the Euclidean distance of far off sensors (i.e., the ones that are not in each other's communication range) are approximated by the shortest path between them. It was recently shown that having local connectivity, the shortest path is a reliable estimate of far off sensors \cite{OKM09}.  Second, to estimate the relative positions of sensors, multidimensional scaling is applied to the approximated distance matrix. However, one can easily see that given faraway sensor's distances, the shortest path is a very coarse estimate of the distance between the close-by sensors. This makes \textsc{Mds-Map} perform very poorly in our setting.

One of the most prominent algorithms for centralized sensor localization is based on semi-definite programming (\textsc{Sdp}). The method was first introduced by Biswas et al. in \cite{SDP06} and solves the sensor localization problem using convex relaxation. From a practical point of view, the major problem of \textsc{Sdp}-based methods is their heavy computations. According to \cite{SDP06}, the sensor localization for more than 200 sensors is computationally prohibitive.  Theoretical guarantees of such methods were provided recently by Javanmard et al. \cite{Javanmard}.  As their results suggest, in the case of sensor localization, once the number of sensors grow, one cannot reduce the error of semidefinite programming below a threshold unless one increases the communication range and hence the power consumption of sensors.  We will show, however, using the matrix completion, the error decreases as the number of transmitter/receivers grows.

In the core of our proposed method is matrix completion,  the problem that aims to recover a low rank matrix from its randomly known entries. It is easy to show that a matrix formed by pairwise distances is low rank (see Lemma \ref{lem:rank3}). Based on this property, Drineas et al. suggested using matrix completion for inferring the unknown distances \cite{Drineas}. However, their analysis relies on the assumption that even for faraway nodes, there is a nonzero probability of communication. This assumption severely restricts  the applicability of their result in practice.  Thinking back to duality between sensor localization and our problem, this assumption suggests that in our case the pairwise distances of nearby transmitters/receivers can be obtained with a nonzero probability, an assumption that does not hold. Fortunately, in the past two years, there has been many improvements on the matrix completion. Cand{\`e}s et al.~showed that a small random fraction of the entries suffices to reconstruct a low rank matrix \textit{exactly.} In a series of papers \cite{KMO09exact,KMO09noise, KOImpl09},  Keshavan et al.~studied an efficient implementation of a matrix completion algorithm so called \opt and showed its optimality. Furthermore, they proved that their algorithm is robust against noise \cite{KMO09noise}. In view of this progress, we were able to show that \opt is also capable of finding the missing nearby distances in our scenario and hence provide us with their corresponding ToFs.  To the best of our knowledge, all the above work as well as the recent matrix completion algorithms \cite{Recht09, Recht11}  only deal with the \textit{random} missing entries. However, in our case, we are encountered with the \textit{structured} missing entries in addition to random ones (see Section \ref{sec:tof_estim}), an aspect that was absent from the previous work. Therefore, one of our contributions is to provide analytic bounds on the error of \opt in the presence of structured missing entries.  

The organization of this paper is as follows; In Section~\ref{sec:mod_def}, we define the model used in circular tomography and introduce the tools used for calibration in such a setup. In Section~\ref{sec:dist_comp}, we present the mathematical basis for the problem. In Sections~\ref{sec:mat_comp} and \ref{sec:pos_recons} an overview of matrix completion and multidimensional scaling methods is provided. Then in Section~\ref{sec:main_res} our main results for calibration are presented. Section~\ref{sec:proofs} contains the proofs for the main results and finally Section~\ref{sec:experimental_results} is devoted to the simulation results. 
\end{section}

%%%%%%%%%%%%%%%%%% Circular Tomography %%%%%%%%%%%%%%%%
\vspace{-0.4cm}
\begin{section}{Circular Time of Flight Tomography}
\label{sec:mod_def}
The focus of this research is ultrasound tomography with circular apertures. In this setup, $n$ ultrasound transmitters and receivers are installed on the interior edge of a circular ring and an object with unknown acoustic characteristics is placed inside the ring. At each time instance a transmitter is fired, sending ultrasound signals with frequencies ranging  from hundreds to thousands of kHz, while the rest of the sensors record the received signals. The same process is repeated for all the transmitters. Each one of $n$ sensors on the ring is capable of transmitting and receiving ultrasound signals. The aim of tomography in general is to use the recorded signals in order to reconstruct the characteristics of the enclosed object (e.g.~sound speed, sound attenuation, etc.). The general configuration for such a tomography device is depicted in Fig.~\ref{fig:circlar_setup}. Employing these measurements, an inverse problem is constructed, whose solution provides the acoustic characteristics of the enclosed object. 

\begin{figure}[tb]
\centering
\includegraphics[scale=1.2]{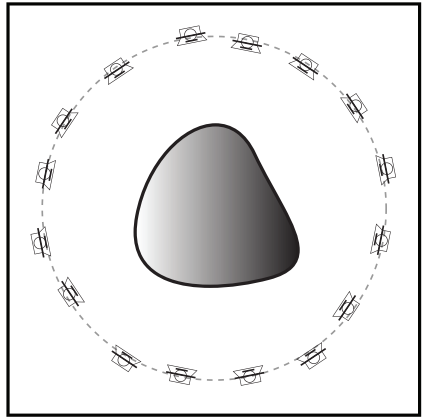}
\caption{Circular setup for ultrasound tomography considered in this work. Ultrasound transducers are distributed on the edge of a circular ring and the object with unknown characteristics is put inside. Transmitters and receivers are collocated. Transducers are fired each in turn while the rest of sensors recording the ultrasound signals reaching them. In practice, the positions deviate from an ideal circle.}
\label{fig:circlar_setup}
\end{figure}

There are two common methods for solving the inverse problem. The solutions are either based on the wave equation~\cite{nat01} or the bent-ray theory~\cite{iva08}. Both techniques consist of forward modeling the problem and comparing the simulation results with the measured data. For the details see \cite{nat01} and \cite{iva08}. Nevertheless, in both cases, in order to simulate the forward model and rely on the recorded data, a very precise estimate of the sensor positions is needed. In most applications (e.g.,~\citep{iva07, sim07, nat08}) it is assumed that the sensors are positioned equidistant apart on a circle and no later calibration is performed to find the exact sensor positions. The main objective of this paper is to estimate the precise positions of the sensors. 

\subsection{Homogeneous Medium and Dimensionality Reduction }
%In order to estimate the sensor positions, we use the bent-ray tomography technique. In this method, the region of interest is descretized and an unknown sound speed $c_i$ is associated to each tile in the region. Once each transmitter is fired and receivers recorded the signals, the relative time-of-flights (ToF) between the pairs of transmitters and receivers will be measured. Afterwards, a non-linear set of equations is constructed as below
%\begin{equation}
%\mathcal{L}(\x)  = \T\,,
%\end{equation}
%where $\mathcal{L}(\x)$ is a nonlinear transform which relates the lengths of the bent-rays from a transmitter to a receiver to the travel time between them. More precisely, the input of this nonlinear transform is the vector $\x$ whose $i$-th entry $x_i$,  is the inverse of the sound speed in the $i$-th tile and the output is the ToFs between each pair of sensors.  

In order to estimate sensor positions, we utilize the ToF measurements for a homogeneous medium (e.g.~water in the context of breast cancer detection).  Let's assume that the mutual ToFs are stored in a matrix $\T$. In a homogeneous medium, entries of $\T$ represent the time travelled by sound in a straight line between each pair of a transmitter and receiver. 

Knowing the temperature and the characteristics of the medium inside the ring, one can accurately estimate the constant sound speed $c_0$. Thus, it is  reasonable to assume that $c_0$ is fixed and known. Having the ToFs for a homogeneous medium where no object is placed inside the ring, we can construct a distance matrix $\D$ consisting of the mutual distances between the sensors as
\begin{equation}
\D = \left[d_{i,j}\right] = c_0 \T\,, \quad \quad \T= \left[t_{i,j}\right]\,, \quad i,j \in \{1,\cdots, n\}
\label{eq:dist_mat}
\end{equation}
where $t_{i,j}$ is the ToF between sensors $i$ and $j$ and $n$ is the total number of sensors around the circular ring. Notice that the only difference between the ToF matrix $\T$, and distance matrix $\D$, is the constant $c_0$. This is why in the sequel our focus will mainly be on the distance matrix rather than the actual measured matrix $\T$.

Since the enclosed medium is homogeneous, the matrix $\T$ is a symmetric matrix with zeros on the diagonal and so is the matrix $\D$. Even though, the distance matrix $\D$ is full rank in general, a simple point-wise transform of its entries will lead to a low rank matrix. More precisely, we can prove (see Appendix~\ref{proof:rank3}) the following lemma:

\begin{Lemma}
\label{lem:rank3}
If one constructs the squared distance matrix $\bD$ as 
\begin{equation*}
\bD = \left[d_{i,j}^2\right]\,,
\end{equation*}
then the matrix $\bD$ has rank at most 4 \cite{dri06} and if the sensors are placed on a circle, the rank is exactly 3.
\end{Lemma}
%\begin{proof}
%The proof for the general case where the sensors are not on a circle is provided in \cite{dri06}. In the circular case however, we have $\bD_{i,j} = \norm{\x_i}^2 + \norm{\x_j}^2 - 2 \x_i^T\x_j = 2r^2 - 2\x_i^T\x_j$, where $r$ is the circle radius. Thus, the squared distance matrix is decomposable to 
%\begin{equation*}
%\bD = \V \bm{\Sigma}\V^T\,,
%\end{equation*}
%where 
%\begin{equation*}
%\V= \begin{bmatrix}
%r & x_{1,1} & x_{1,2}\\
%\vdots & \vdots & \vdots\\
%r & x_{n,1} & x_{n,2}
%\end{bmatrix} \,, \quad \bm{\Sigma}  = \begin{bmatrix}
%2 & 0 & 0\\
%0 & -2 & 0\\
%0 & 0 & -2
%\end{bmatrix}\,.
%\end{equation*}
%\end{proof}
In practice, as we will explain in the next section, many of the the entries of the ToF matrix (or equivalently  the distance matrix) are missing and there is an unknown time delay added to all the measurements.   
\vspace{-.4cm}
\subsection{Time of Flight Estimation}
\label{sec:tof_estim}
Several methods for ToF estimation have been proposed in the signal processing community \cite{iva08, li09}. These methods are also known as time-delay estimation in acoustics~\citep{hua04}. In these methods, the received signal is compared to a reference signal (ideally the sent signal), and the relative delay between the two signals is estimated. Since the sent signal is not available in most cases, the received signal through the object is compared to the received signal when the underlying medium is homogeneous. However, this assumption is not true in our case. In the calibration phase, we have only signals passed though the homogeneous medium. Thus, there is not any reference signal to find the relative time-of-flights. 

Because of the above limitations, we are forced to estimate the \textit{absolute} ToFs. For this purpose, we use the first arrival method. This method probes the received signal and defines the time-of-flight as the time instant at which the received signal power exceeds a predefined threshold. 

In practical screening systems, to record measurements for one fired transmitter, all the sensors are turned on simultaneously and after some unknown transition time (which is caused by the system structure, different sensor responses, etc.), the transmitter is fed with the electrical signal and the receivers start recording the signal. This unknown time may change for each pair of transmitters and receivers. We will see that this unknown time delay plays an important role in sensor position estimation.

The beam width of the transducers and the transition behaviour of the ultrasonic sensors prevent the sensors to have a reliable ToF measurement  for close-by neighbours. This results in incorrect  ToF values for the sensors positioned close to each other. Therefore, numbering the sensors on the ring from 1 to $n$, in the ToF matrix $\T$, we will not have measurements on a certain band around the main diagonal and on the lower left and upper right parts as well. We call these missing entries as \textit{structured missing entries}. This is illustrated in Fig.~\ref{fig:struc_miss_ent}. The links shown by dashed lines do not contribute in the ToF measurements, because the beam for the transmitter does not cover the gray part. 
%One can define a circle around each transmitter for which the sensors lying inside do not have correct ToF values. 

\begin{figure}[tb]
\centering
\includegraphics[scale=1.5]{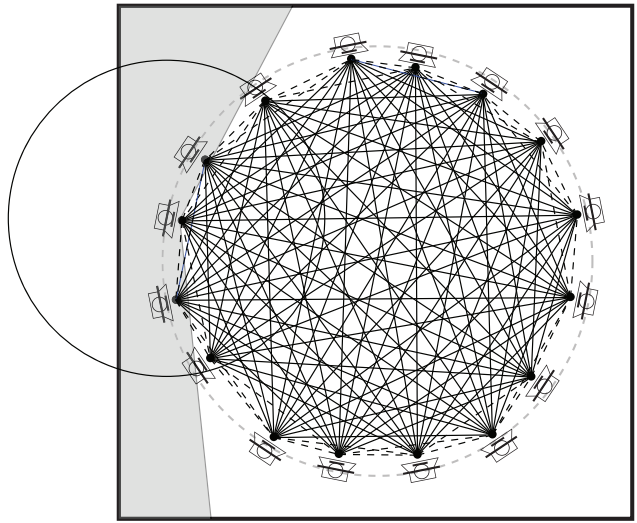}
\caption{The beam width of the transmitter causes the neighbouring sensors not to have reliable ToF measurements. This is shown by dashed lines in the figure. The area shown in gray corresponds to the part which is not covered by the transmitter's wave beam. This results in the structured missing entries.}
\label{fig:struc_miss_ent}
\end{figure}

During the measurement procedure, it may also happen that some sensors do not act properly and give outliers. Thus, one can perform a post processing on the measurements, in which a smoothness criterion is defined and the measurements not satisfying this criterion are removed from the ToF matrix. We address these entries as \textit{random missing entries}. An instance of the ToF matrix with the structured and random effects is shown in Fig.~\ref{fig:tof_missing_entries}, where $\T_{_{\text{inc}}}$ denotes the incomplete ToF matrix and the gray entries correspond to the missing entries.
\begin{figure}[tb]
\centering
\psfrag{T}{$\T_{_{\text{inc}}}$}
\psfrag{=}{$\,=$}
\psfrag{?}{\footnotesize{$\;$}}
\includegraphics[scale=1]{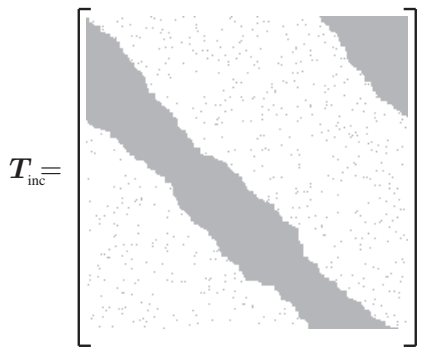}$\,.$
\caption{A sample incomplete ToF matrix with structured and random missing entries.}
\label{fig:tof_missing_entries}\end{figure}
Furthermore, in practice, the measurements are corrupted by noise. 

The above mentioned problems result in an incomplete and noisy matrix $\T$, which cannot be used for position reconstruction, unless the time delay effect is removed,  the unknown entries are estimated, and the noise is smoothed. 
\end{section}

%%%%%%%%%%%%%%%%%%% Problem Setting %%%%%%%%%%%%%%%%%%%
\vspace{-0.4cm}
\begin{section}{Problem Setting}
\label{sec:dist_comp}
We observed that the distance matrix, when the aperture is in homogeneous medium, is calculated as in \eqref{eq:dist_mat}. We also saw in the previous section that the measurements for the ToF matrix $\T$ have three major problems : they are noisy, some of them are missing, and the measurements contain some unknown time delay. For simplicity, we will assume that this time delay is constant for all the transmitters, namely all the transmitters send the electrical signal after some fixed but unknown delay $t_0$. Hence, we can rewrite the ToF matrix as follows
\begin{equation*}
\tT = \T + t_0\A + \Z_0\,,  
\end{equation*} 
where $\T$ consists of ideal measurements for ToF, $\Z_0$ is the noise matrix and $\A$ is defined as
\begin{equation*}
\A = \left[a_{i,j}\right], \quad a_{i,j} =  
\begin{cases}
1 & \text{if } i \neq j\,,\\
0 & \text{otherwise}\,.
\end{cases}
\end{equation*} 
With the above considerations, the distance matrix can also be written as
\begin{equation}
\tD = \D + d_0 \A + \Z\,,
\label{eq:Dt_Def}
\end{equation}
where $\D = c_0 \T$, $d_0=c_0 t_0$, and $\Z=c_0\Z_0$.

In our model we do not assume that the sensors are placed exactly on the ring. What happens in practice is that the sensor positions deviate from the circumference and our ultimate goal is to estimate these deviations or equivalently the correct positions (see Fig.~\ref{model}). The general positions taken by sensors are denoted by the set of vectors $\{\x_1,\ldots,\x_n\}$. 
\begin{figure}[tb]
\centering
\psfrag{a}{$r$}
\psfrag{c}{$\delta_n$}
\psfrag{b}{$a$}
\includegraphics[scale=0.7]{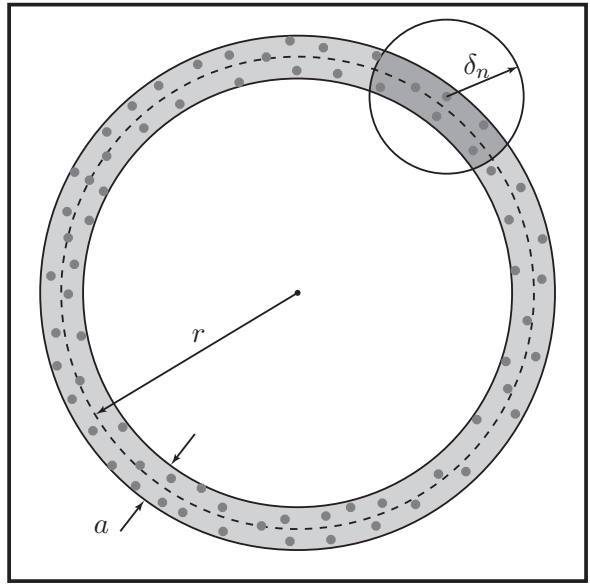}
\caption{Sensors are distributed around a circle of radius $r$ with small deviations from the circumference.}
\label{model}
\end{figure}

As described earlier, there are two contributions to missing entries. 
One is the missing measurements of close-by sensors, 
which we call  structured missing entries. 
The other is the missing measurements due to 
random malfunction of sensors, 
which we call random missing entries. First, to incorporate the structured missing entries, 
we assume that any measurements between sensors of 
distance less than $\delta_n$ are missing (see Figure \ref{model}). Hence, the number of structured missing entries depends on $\delta_n^2$. We are interested in the regime where we have a small 
number of structured missing entries per row in the large systems limit.
Accordingly, typical range of $\delta_n$ of interest is $\delta_n=\Theta(\,r\sqrt{\log n/n})$.
A random set of structured missing indices $S\subseteq [n]\times[n]$ is 
defined from $\{\x_i\}$ and $\delta_n$, by 
\begin{eqnarray*}
	S = \{(i,j):\; d_{i,j} \leq \delta_n \text{ and } i\neq j\}\;, \label{eq:def_S}
\end{eqnarray*}
where $d_{i,j}=\norm{\x_i-\x_j}$. 
Then, the structured missing entries are denoted by a matrix
\begin{eqnarray*}
\label{eq:def_Dbs}
	\Ds_{i,j} = 
		\begin{cases}
			\D_{i,j} & \text{if } (i,j) \in S \;,\\
			0 & \text{otherwise} \;.
		\end{cases}
\end{eqnarray*}
Note that the matrix $\Dsb=\D-\Ds$ captures the noiseless distance measurements 
that is not effected by structured missing entries. This way, we can interpret the matrix $\Ds$ as additive noise in our model. Likewise, for the constant additive time delay we can define
\begin{eqnarray*}
\label{eq:def_Asb}
	\Asb_{i,j} = 
		\begin{cases}
			\A_{i,j} & \text{if } (i,j) \in S^\perp \;,\\
			0 & \text{otherwise} \;,
		\end{cases}
\end{eqnarray*}
where $S^\perp$ denotes the complementary set of $S$. 
Next, to model the noise we add a random noise matrix $\Zsb$. 
\begin{eqnarray*}
\label{eq:def_Zsb}
	\Zsb_{i,j} = 
		\begin{cases}
			\Z_{i,j} & \text{if } (i,j) \in S^\perp \;,\\
			0 & \text{otherwise} \;.
		\end{cases}
\end{eqnarray*}
We do not assume a prior distribution on $\Z$, and 
the main theorem is stated for any general noise matrix $\Z$, 
deterministic or random.
One practical example of $\Z$ is an i.i.d. Gaussian model.

Finally, to model the random missing entries, 
we assume that each entry of $\Dsb +t_0c_0\Asb + \Zsb$ is 
sampled with probability $p_n$. In the calibration data, 
we typically see a small number of random missing entries.
Hence, in order to model it we assume that $p_n=\Theta(1)$.
Let $E\subseteq [n]\times[n]$ denote the subset of indices 
which are not erased by random missing entries. 
Then a projection $\cP_E:\R^{n \times n} \to \R^{n\times n}$ is defined as 
\begin{eqnarray*}
\label{eq:def_cP}
 \cP_E(\M)_{i,j} = 
	\begin{cases}
		\M_{i,j} & \text{if } (i,j) \in E\;,\\
		0 & \text{otherwise}\;.
	\end{cases}
\end{eqnarray*}
We denote the observed measurement matrix by 
\begin{equation}
\label{measurement}
\N^E=\cP_E(\Dsb +d_0\Asb + \Zsb),
\end{equation}
where $d_0=t_0c_0$ is a constant. Notice that the matrix $\N^E$ has the same shape as $\T_{_{\text{inc}}}$ shown already schematically in Fig.~\ref{fig:tof_missing_entries}. Now we can state the goal of our calibration problem:
 
 Given the observed matrix $\N^E$ and  
the missing indices $S\cup E^\perp$, 
we want to estimate a matrix $\hD$ which is close to the correct distance matrix $\D$. Then by using $\hD$ we would like to estimate the sensor positions. 

In order to achieve this goal, there are two obstacles we need to overcome. First, we need to estimate the missing entries of $\N^E$ and second, we want to find the sensor positions given approximate pairwise distances. The former is done by employing a matrix completion algorithm and the latter by using the multidimensional scaling method. 
\end{section}

%%%%%%%%%%%%%%%%% Matrix Completion %%%%%%%%%%%%%%%%%%%
\vspace{-0.4cm}
\begin{section}{Matrix Completion}
\label{sec:mat_comp}
{\sc OptSpace}, introduced in \cite{KMO09exact}, is an algorithm 
for recovering a low-rank matrix from noisy data with missing entries. The steps are shown in Algorithm \ref{algo:optspace}. Let $\M$ be a rank-$q$ matrix of dimensions $n\times n$, $\Z$ the measurement noise, 
and $E$ the set of indices of the measured entries. Then, the measured noisy and incomplete matrix is 
$\M^E=\cP_E(\M+\Z)$.

\begin{algorithm} 
\caption{{\sc OptSpace} \cite{KMO09exact}} 
\label{algo:optspace} 
\begin{algorithmic}[1]
\REQUIRE Observed matrix $\M^E=\cP_E(\M+\Z)$.
\ENSURE Estimate $\M$.
\STATE  Trimming: remove over-represented columns/rows;
\STATE Rank-$q$ projection on the space of rank-$q$ matrices according to \eqref{q-projection};
\STATE Gradient descent: Minimize a cost function $F(\cdot)$ defined in \cite{KMO09exact};
\end{algorithmic} 
\end{algorithm}

In the trimming step, a row or a column is over-represented if 
it contains more samples than twice the average number of samples 
per row or column.
These rows or columns can dominate 
the spectral characteristics of the observed matrix $\M^E$. Thus, some of their entries are removed uniformly at random from the observed matrix.
Let $\tM^E$ be the resulting matrix of this trimming step.
This trimming step is presented here for completeness, 
but in the case when $p_n$ is larger than some fixed constant (like in our case where $p_n=\Theta(p)$), 
$\M^E$=$\tM^E$ with high probability and the trimming step can be omitted.

In the second step, we first compute the singular value decomposition (SVD)
of $\tM^E$. 
\begin{equation*}
\tM^E = \sum_{i=1}^{n}\sigma_i(\tM^E)  u_iv_i^T\, ,
\end{equation*}
where $\sigma_i(\cdot)$ denotes the $i$-th singular value of a matrix.
Then, the rank-$q$ projection returns the matrix 
\begin{equation}
\label{q-projection}
\cP_q(\tM^E) =  
(1/p_n)\sum_{i=1}^{q}\sigma_i(\tM^E)u_iv_i^T,
\end{equation}
obtained by setting to $0$ all but the $q$ largest singular values.

Starting from the initial guess provided by the rank-$q$ projection $\cP_q(\tM^E)$, 
the final step solves a minimization problem stated as the following \cite{KMO09exact}: \\
Given $\X \in \R^{n\times q}, \Y \in \R^{n\times q}$ with $\X^T\X = \bm{1}$ and $\Y^T\Y = \bm{1}$, define
\begin{equation*}
F(\X,\Y) = \min_{\S\in\R^{q\times q}} \mathcal{F}(\X,\Y,\S)\,,
\end{equation*}
\begin{equation*}
\mathcal{F}(\X,\Y,\S) = \frac{1}{2}\sum_{(i,j)\in E} (\M_{i,j}- (\X\S\Y^T)_{i,j})^2\,.
\end{equation*}
Values for $\X$ and $\Y$ are computed by minimizing  $F(\X,\Y)$. This consists of writing $\cP_q(\tM^E) =  \X_0\S_0Y_0^T$ and minimizing $F(\X,\Y)$ locally with initial condition $\X = \X_0$ and $\Y = \Y_0$.
 This last step tries to get us as close as possible to the correct low rank matrix $\M$. 
\end{section}

%%%%%%%%%%%%%%%%%%%% Position Reconstruction %%%%%%%%%%%%%%
\vspace{-0.4cm}
\begin{section}{Position Reconstruction}
\label{sec:pos_recons}

Even if we had a good estimate of $\D$, how we would position the sensors is not a trivial question. Multidimensional scaling (MDS) is a technique used in finding the configuration of objects in a low dimensional space
such that the measured pairwise distances are preserved.
If all the pairwise distances are measured without error, then 
a naive application of MDS exactly recovers the configuration of sensors \cite{MDS,SRZ03,dri06}.

\begin{algorithm} 
\caption{Classical Metric MDS \cite{SRZ03}.} 
\label{alg:MDS} 
\begin{algorithmic}[1]
\REQUIRE Dimension $\eta$, estimated  squared distance matrix $\bD$
\ENSURE Estimated positions $\MDS_{\eta}(\bD)$
\STATE Compute $(-1/2)\L \bD \L$, where $\L=\ind_n-(1/n)\ones_n\ones_n^T$;
\STATE Compute the best rank-$d$ approximation $\U_{\eta}\Si_{\eta} \U_{\eta}^T$ of $(-1/2)\L \bD\L$;
\STATE Return $\MDS_{\eta}(\bD)\equiv \U_{\eta}\Si_{\eta}^{1/2} $.
\end{algorithmic} 
\end{algorithm}

% Formal definition of MDS
There are various types of MDS techniques, but, throughout this paper, 
by MDS we refer to the classical metric MDS, which is defined as follows.
Let $\L$ be an $n \times n$ symmetric matrix such that 
\begin{equation}
\label{matrixL}
\L=\ind_n-(1/n)\ones_n\ones_n^T,
\end{equation}
where $\ones_n \in \R^n$ is the all ones vector and 
$\ind_n$ is the $n \times n$ identity matrix.
Let ${\sf MDS}_{\eta}(\bD)$ denote the $n\times {\eta}$ matrix 
returned by MDS when applied to the squared distance matrix $\bD$. The task is to embed $n$ objects in a ${\eta}$ dimensional space $\R^{\eta}$. In our case for instance, where we want to find the position of sensors on a two dimensional space, we have ${\eta}=2$.
%Then the classical metric MDS of an $n\times n$ squared distance matrix $D$
%computes the first $d$ principal components of $(-1/2)\Ln D\Ln$. 
Then, in the equation, given the singular value decomposition (SVD) 
of a symmetric and positive semidefinite matrix $(-1/2)\L \bD \L$ 
as $(-1/2)\L \bD \L = \U\Si \U^T$, 
\begin{eqnarray*}
 \MDS_{\eta}(\bD) \equiv \U_{\eta} \Si_{\eta}^{1/2}\;,
\end{eqnarray*}
where $\U_{\eta}$ denotes the $n\times {\eta}$ left singular matrix 
corresponding to the ${\eta}$ largest singular values and $\Si_{\eta}$
denotes the ${\eta}\times {\eta}$ diagonal matrix with ${\eta}$ largest singular values in the diagonal.
This is also known as the {\sc MDSLocalize} algorithm in \cite{dri06}.
Note that since the columns of $\U$ are orthogonal to $\ones_n$ by construction, 
it follow that 
\begin{equation}\label{LMDS}
\L\cdot\MDS_{\eta}(\bD) = \MDS_{\eta}(\bD).
\end{equation}

%In other words, $\hX\hX^T$ is the rank-$d$ projection of $(-1/2)\Ln D\Ln$,
%where we define the rank-$d$ projection of a matrix $A$ as  
%$\cP_d(A)=\sum_{i=1}^d{\sigma_iu_iv_i^T}$.
%Here, $u_i$ and $v_i$ are the left and right singular vectors of $A$, respectively, 
%corresponding to the $i$th singular value $\sigma_i$.

%Property of MDS
It can be easily shown that 
when MDS is applied to the correct and complete squared distance matrix without noise, 
the configuration of sensors are exactly recovered \cite{dri06}.
This follows from
\begin{equation}
 -\frac{1}{2}\L \bD \L = \L \X\X^T\L\;, 
 \label{LMDS2}
\end{equation}
where $\X$ denotes the $n\times {\eta}$ position matrix in which the $i$-th row corresponds to $\mathbi{x}_i$, the ${\eta}$ dimensional  position vector of sensor $i$. Note that we only get the configuration and not the absolute positions, 
in the sense that $\MDS_{\eta}(\bD)$ is one version of infinitely many solutions 
that matches the distance measurements $\D$. 
%which are equal up to rigid transformation (rotations, reflections and translations). 
%Multiplicity
%The squared distance matrix $D$ is a function of the sensor positions $X$
%such that $D(X)_{ij}=||x_{i}-x_{j}||^2$, where $x_i$ and $x_j$ denote 
%the $i$th and $j$th rows of $X$, respectively.
Intuitively, it is clear that the pairwise distances are invariant to 
a rigid transformation (a combination of 
rotation, reflection and translation) of the positions $\X$, 
and therefore there are multiple instances of $\X$ that result in the same $\D$.
For future use, we introduce a formal definition of rigid transformation 
and related terms.

Denote by $\Orth({\eta})$ the group of orthogonal ${\eta}\times {\eta}$ matrices.
A set of sensor positions $\Y\in\R^{n\times {\eta}}$ is a rigid transform of $\X$, 
if there exists a ${\eta}$-dimensional shift vector $\bm{s}$ 
and an orthogonal matrix $\Q\in\Orth({\eta})$ such that 
\begin{eqnarray*}
 \Y = \X\Q+\ones_n \bm{s}^T \;.
\end{eqnarray*}
$\Y$ should be interpreted as 
a result of first rotating (and/or reflecting) sensors in position $\X$ by $\Q$ and then 
adding a shift by $\bm{s}$. 
%Note that we could have shifted the 
%sensors by another shift vector $s'$ and then rotated it 
%to get the same positions $Y=(X+\ones_n s'^T)Q$.
Similarly, when we say two position matrices $\X$ and $\Y$ 
are equal up to a rigid transformation, 
we mean that there exists a rotation $\Q$ and a shift $\bm{s}$ such that 
$\Y = \X\Q+\ones_n \bm{s}^T$.
Also, 
%With this definition of rigid transformation, 
%we can define what it means for a function to be invariant under rigid transformation. 
we say a function $f(\X)$ is invariant under rigid transformation 
if and only if for all $\X$ and $\Y$ that are equal up to a 
rigid transformation, we have $f(\X)=f(\Y)$.
Under these definitions, it is clear that $\D$ 
is invariant under rigid transformation, since for all $(i,j)$, 
$ \D_{ij} = \norm{\x_i-\x_j} = \norm{(\x_i\Q+\bm{s}^T)-(\x_j\Q+\bm{s}^T)} \;$, 
for any $\Q\in\Orth({\eta})$ and $\bm{s}\in\R^{\eta}$. 

Let $\hX$ denote an $n \times {\eta}$ estimation for $\X$ 
with estimated position for sensor $i$ in the $i$-th row. 
Then, we need to define a metric for the distance 
between the original position matrix $\X$ 
and the estimation $\hX$ which is invariant under 
rigid transformation of $\X$ or $\hX$. 

The matrix $\L$ defined in \eqref{matrixL} is a symmetric matrix with rank $n-1$
which eliminates the contributions of the translation. More precisely,
\begin{displaymath}
\L\X=\L(\X+\ones \bm{s}^T),
\end{displaymath} 
for all $\bm{s}\in R^{\eta}$.  
We can show that $\L$ has the following properties.
\begin{Lemma}\cite{SRZ03,dri06,OKM09}
\label{Lprop}
Let the matrix $\L$ be defined as in \eqref{matrixL}. Moreover, let $\X$ and $\hX$ be two position matrices with dimension $n\times {\eta}$. Then, we can show that   
\begin{itemize}
\item $\L \X\X^T\L$ { is invariant under rigid transformation}. 
\item $\L \X\X^T\L=\L \hX\hX^T\L$ { implies that $\X$ and $\hX$ are equal up to a rigid transformation}. 
\end{itemize}
\end{Lemma}
This naturally defines the following distance between $\X$ and $\hX$.
\begin{eqnarray}
\label{distance}
	d(\X,\hX) = \frac{1}{n}\norm{\L \X\X^T\L-\L\hX\hX^T\L }_F \;,
\end{eqnarray}
where $\norm{\cdot}_F$ denotes the Frobenius norm.

According to Lemma~\ref{Lprop}, this distance is invariant to rigid transformation of $\X$ and $\hX$. 
Furthermore, $d(\X,\hX)=0$ implies that $\X$ and $\hX$ 
are equal up to a rigid transformation.
We later state our theoretical results in terms of the distance defined in \eqref{distance}.
\end{section}

%%%%%%%%%%%%%%%%%%%% Main Results %%%%%%%%%%%%%%%%%%
\vspace{-0.4cm}
\begin{section}{Main results}
\label{sec:main_res}
\begin{table}[tb]
\caption{Summary of Notation.}
\ra{0.8}
\centering
\begin{tabular}{@{}ll | ll@{}}
\toprule
Symbol & Meaning & Symbol & Meaning\\
\midrule
$n$ & number of sensors & $\D$ & complete noiseless distance matrix\\
$r_0$ & radius of the circle from which the sensors deviate & $\bD$ & squared distance matrix\\
$a/2$ & maximum radial deviation from the circle & $\tD$ & noisy distance matrix\\	
$\cP_E$ & projection into matrices with entries on index set $E$ & $\hbD$ & estimated squared distance matrix\\
$t_0$ & unknown time delay added to the ToF measurements & $\Z$ & noise matrix\\
$d_0$ & distance mismatch caused by the unknown time delay & $\N^E$ & observed matrix\\
$p_n$ & probability of having random missing entries & $\X$ &positions matrix\\
$\delta_n$ & radius of the circle defining structured missing entries & $\hX$ &estimated positions matrix\\
$\Ds$ & distance matrix with observed entries on index set $S$ & &\\
\bottomrule
\end{tabular}
\label{tab:notations}
\end{table}
We saw that the \opt algorithm is not directly applicable to the squared distance matrix because of the unknown delay. Since $\A$ in \eqref{eq:Dt_Def} is a full rank matrix, the matrix $\tD\odot \tD = [{\tilde{d}_{i,j}}^2]$ no longer has rank four. Moreover, since the measurements are noisy, one cannot hope for estimating the exact value for $d_0$. Therefore, in the following we will provide error bounds on the reconstruction of the positions assuming that the time delay (equivalently $d_0$) is known. Afterwards, a heuristic method is proposed to estimate the value of $d_0$. 

In Table \ref{tab:notations} the set of important notations used in the sequel is summarized. 

%\textbf{CAREFULL:} It is also important to note that in the following results, to be able to prove the validity of the bounds we have assumed that the sensors are distributed on a circle (and $\bD$ has rank three). In the case that they are not on a perfect circle, the rank will increase to four, and we will show by simulations that the results are still valid and the problem is stable. 

\begin{thm}
\label{thm:main1}
Assume $n$ sensors are distributed independently and uniformly at random on a circular ring of width $a$ with central radius $r_0$ as in Fig.~\ref{fig:struc_miss_ent}. The resulting distance matrix $\D$ is corrupted by structured missing entries $\Ds$ and measurement noise $\Zsb$. Further, the entries are missing randomly with probability $p_n$. Let $N^E = \P_E(\D - \Ds+ \Zsb)$ denote the observed matrix. Define $\bD$ as the squared distance matrix. Assume $\delta_n = \delta \,r_0 \,\sqrt{\log n/n}$ and $p_n = p$. Then, there exist constants $C_1$ and $C_2$, such that the output of \opt $\widehat{\bD}$ achieves
\begin{equation}
\label{eq:mainthm1}
\frac{1}{n}\|{\bD - \widehat{\bD}}\|_{\text{F}} \leq C_1 \left(\sqrt{\frac{\log n}{n}}\right)^3 + C_2 \frac{\norm{\P_E(\Ysb)}_2}{p\,n}\,,
\end{equation}
with probability larger than $1 - n^{-3}$, provided that the right hand side is less than $\sigma_4(\bD)/n$. We have $\Ysb_{i,j}  = {\Zsb}^2_{i,j}+ 2\Zsb_{i,j}\Dsb_{i,j}$. % and 
%\begin{equation*}
%C_1 = c\,\delta^3\,(r_0+a)^2\,.\end{equation*}
\end{thm}

The above theorem, in great generality, holds for 
any noise matrix $\Z$, deterministic or random. 
%However, to make sense of the above results 
%it is convenient to consider i.i.d. Gaussian noise.
%The following theorem provides an upper bound 
%on the operator norm $||\cP_E(\Zsb)||_2$ in this typical case. 
%
%\begin{thm}
%\label{thm:main3}
%Assume the entries of the noise matrix $Z$ are independent random variables 
%following a Gaussian distribution with zero mean and variance $\sigma^2$.
%Under the same hypothesis as Theorem \ref{thm:main1}, 
%there exists a numerical constant $C$ such that, 
%\begin{eqnarray*}
%	||\cP_E(\Zsb)||_2 \leq C \sigma \sqrt{\delta\,p\,n\,\log n} \;,
%\end{eqnarray*}
%with probability larger than $1-1/n^3$.
%\end{thm}
%
%Note that in the case of i.i.d. Gaussian noise with finite variance, 
%the right-hand side of Eq.~(\ref{eq:thm2_main}) 
%is proportional to $\sqrt{\log n/n}$.
%This implies that for $n$ large enough, 
%Eq.~(\ref{eq:thm2_main}) is smaller than $(3/16)r$.
The above guarantees only hold `up to numerical constants'. 
To see how good {\sc OptSpace} is in practice, 
we need to run numerical experiments.
For more results supporting the robustness of {\sc OptSpace}, 
we refer to \cite{KOImpl09}.

\begin{corollary}
\label{thm:main2}
Applying multidimensional scaling algorithm on $\hbD$, the error on the resulting coordinates will be bounded as follows
\begin{equation}
	d( \X , \hX) \leq C_1 \left(\sqrt{\frac{\log n}{n}}\right)^3 + C_2 \frac{\norm{\P_E(\Ysb)}_2}{p\,n}\,, 
\end{equation}
with probability larger than $1-1/n^3$. (The proof is given in Appendix~\ref{proof:thm2})
\end{corollary}

In Algorithm~\ref{alg:find_t_0}, we propose a heuristic method for estimating the value of $d_0$ along with completion of the squared distance matrix.
%However, we can prove the following theorem. 

\begin{algorithm}[tb]
\caption{Finding $d_0$.} 
\label{alg:find_t_0} 
\begin{algorithmic}[1]
\REQUIRE Matrix $\N^E$;
\ENSURE Estimate $d_0$;
%\STATE Set $\N^E_{(0)} = \N^E$;
\STATE Construct the candidate set $\mathcal{C}_d = \{d_0^{(1)},\ldots,d_0^{(M)}\}$ containing discrete values for $d_0$.
\FOR{$k=1$ to $M$}
\STATE Set $\N^E_{(k)} = \N^E - d_0^{(k)} A^E$; 
\STATE Set $\bN^E_{(k)}=\N^E_{(k)}\odot\N^E_{(k)}$;
\STATE Apply \opt on $\bN^E_{(k)}$ and call the output $\hat{\N}^{(k)}$;
\STATE Apply MDS and let $\X^{(k)}={\MDS}_2(\hat{\N}^{(k)})$;
\STATE Find $c^{(k)}$ \\ 
\hspace{0.3cm}$c^{(k)} = \sum_{(i,j)\in E\cap S^\perp}{\big(d_0^{(k)}+\|\X^{(k)}_i-\X^{(k)}_j\|-\N^E_{i,j}\big)^2}$;
%\STATE \hspace{0.1cm} Set $\N^E_{(k+1)} = \N^E - d_0^{(k)} A^E $;
\ENDFOR
\STATE \hspace{0.1cm} Find $d_0$ satisfying\\
\hspace{0.3cm}$d_0 = d_0^{(l)}, \quad l = \argmin_{k} c^{(k)}$;
\end{algorithmic} 
\end{algorithm}
In fact, this algorithm guarantees that after removing the effect of the time delay, we have found the best rank 4 approximation of the distance squared matrix. In other words, if we remove exactly the mismatch $d_0$, we will have an incomplete version of a rank 4 matrix and after reconstruction, the measured values will be close to the reconstructed ones.
\end{section}

%%%%%%%%%%%%%%%%%%%%% Proofs %%%%%%%%%%%%%%%%%%
\vspace{-0.4cm}
\begin{section}{Proof of Theorem \ref{thm:main1}}
\label{sec:proofs}
This section is dedicated to the proof of our main result. To do so  we apply Theorem 1.2 of \cite{KMO09noise} 
to the rank-$4$ matrix $\bD$ and the observed matrix $\N^E=\cP_E(\bD-\bD^s+\Zsb)$. 

First, we provide the definition of a crucial property of 
$\bD$ which is called {\em incoherence}. 
Following the definition in \cite{KMO09noise}, 
a rank-$4$ symmetric matrix $\bD\in\R^{n \times n}$ is said to be $\mu$-{\em incoherent} 
if the following conditions hold. 
Let $U\Sigma U^T$ be the singular value decomposition of $\bD$. 
\begin{itemize} 

\item[{\bf A0.}] For all  $i\in [n]$, we have  
           $\sum_{k=1}^{4}{U_{i,k}^2} \le 4\mu /n$.
\item[{\bf A1.}] For all $i\in[n]$, $j\in [n]$,
we have $\big| \bD_{i,j}/\sigma_1(\bD) \big|\leq\sqrt{4} \mu/n$.
\end{itemize}
The extra $1/n$ terms in the right hand side are due to the fact that, 
in this paper, we assume that the singular vectors 
are normalized to unit norm, 
whereas in \cite{KMO09noise} the singular 
vectors are normalized to have norm $\sqrt{n}$.

Theorem 1.2 of \cite{KMO09noise} states that if a rank-$4$ matrix $\bD$ is 
$\mu$-{\em incoherent} then 
the following is true with probability at least $1-1/n^3$.
Let $\sigma_i(\bD)$ be the $i$th singular value of $\bD$ and 
$\kappa(\bD)=\sigma_1(\bD)/\sigma_4(\bD)$ be the condition number of $\bD$.
Also, let $\hbD$ denote the estimation returned by {\sc OptSpace} 
with input $N^E=\cP_E(\bD-\bD^s+\Ysb)$. 
Then, there exists numerical constants $C_1$ and $C_2$ such that 
\begin{eqnarray}
	\frac{1}{n}|| \bD - \hbD ||_F \leq C_1\,\frac{\|\cP_E(\bD^s)\|_2 + \|\cP_E(\Ysb)\|_2}{p\,n}\;, \label{eq:thm1proof1}
\end{eqnarray}
provided that 
\begin{eqnarray}
	np\geq C_2\mu^2\kappa(\bD)^6\log n \;,	\label{eq:thm1proof2}
\end{eqnarray}
and 
\begin{eqnarray}
	C_1\,\frac{\|\cP_E(\bD^s)\|_2 + \|\cP_E(\Ysb)\|_2}{p\,n} \leq \frac{\sigma_4(\bD)}{n} \;.	\label{eq:thm1proof3}
\end{eqnarray}
First, using Lemma \ref{lem:2norm}, we show that the bound in \eqref{eq:thm1proof1} 
gives the desired bound in the theorem. 
Then, it is enough to show that there exists a numerical constant $N$ such that 
the conditions in \eqref{eq:thm1proof2} and 
\eqref{eq:thm1proof3} are satisfied with high probability for $n\geq N$. 

% 
% ======================================================================================= 
% 
\begin{Lemma}
\label{lem:2norm}
In the model defined in the previous section, 
$n$ sensors are distributed independently and 
uniformly at random on a circular ring of width $a$ with central radius $r_0$. 
Then, with probability larger than $1-n^{-3}$, there exists a constant $c$ such that 
\begin{equation}
\|\cP_E(\bD^s)\|_2 \leq c\delta^3(r_0+a)^2\Big(\sqrt{\frac{\log n}{n}}\Big)^3p\,n\,.
%\|\cP_E(\bD^s)\|_2 \leq \frac{C \delta^3 \,r^2 \,p\,(\log n)^3}{n^2} \,,
\end{equation}
where $\cP_E(\cdot)$ and $\bD^s$ are defined as in \eqref{eq:def_cP}. The proof of this lemma can be found in Appendix~\ref{proof:2norm}
\end{Lemma}
Now, to show that \eqref{eq:thm1proof2} holds with high probability 
for $n\geq C\log n/p$ for some constant $C$, 
we show that $\kappa\leq f_{\kappa}(r_0, a)$ and $\mu\leq f_{\mu}(r_0,a)$ with high probability, where $f_{\kappa}$ and $f_{\mu}$ are independent of $n$.
Recall that $\kappa(\bD) = \sigma_1(\bD)/\sigma_4(\bD)$. 
We have 
\begin{equation*}
\begin{aligned}
\bD_{i,j}&=\|\x_i\|^2+\|\x_j\|^2-2\x_i^T\x_j \\
&= (r_0+\rho_i)^2+(r_0+\rho_j)^2 - 2\x_i^T\x_j\\ 
&= 2r_0^2 + (2r_0\rho_i+\rho_i^2) + (2r_0\rho_j+\rho_j^2) - 2\x_i^T\x_j\,,
\end{aligned}
\end{equation*}
where $\rho_i$ is distributed in such a way that we have uniform distribution over the circular band. Thus, one can show that  
\begin{equation*}
	\bD = \A\S\A^T\;,
\end{equation*}
where 
\begin{equation*}
	\A = \begin{bmatrix}r_0&x_{1,1}&x_{1,2}&2r_0\rho_1+\rho_1^2\\ \vdots&\vdots&\vdots & \vdots\\ r_0&x_{n,2}&x_{n,2} &2r_0\rho_n+\rho_n^2 \end{bmatrix}, \quad
	\S = \begin{bmatrix}2&0&0&\frac{1}{r_0}\\0&-2&0&0\\0&0&-2&0\\\frac{1}{r_0}&0&0&0\end{bmatrix}\,.
\end{equation*}
One can write $\S$ as
\begin{equation*}
\S = \bm{U}\bm{\Lambda}\bm{U}^{-1}, \quad \bm{\Lambda} = \text{diag}\left(-2, -2, \frac{r_0+\sqrt{1+r_0^2}}{r_0}, \frac{r_0-\sqrt{1+r_0^2}}{r_0}\right)\,,
\end{equation*}
%\begin{equation*}
%\S = \bm{U}\bm{\Lambda}\bm{U}^{-1}, \quad \bm{\Lambda} = \begin{bmatrix} -2&0&0&0\\0&-2&0&0\\0&0&\frac{r_0+\sqrt{1+r_0^2}}{r_0}&0\\0&0&0&\frac{r_0-\sqrt{1+r_0^2}}{r_0}\end{bmatrix}\,,
%\end{equation*}
It follows that $\sigma_1(\bD)\leq \frac{r_0+\sqrt{1+r_0^2}}{r_0}\sigma_1(\A\A^T)$ and $\sigma_4(\bD)\geq \min\Big(2,\frac{\sqrt{1+r_0^2} - r_0}{r_0}\Big)\sigma_4(\A\A^T)$.
We can compute the expectation of this matrix over the distribution of node positions. Having uniform distribution of the sensors over the circular ring, we have for the probability distribution of $\rho$:
\begin{equation*}
p_{\rho}(\rho) = \frac{r_0+\rho}{r_0a},\; \text{for } -\frac{a}{2}\leq\rho\leq \frac{a}{2}\,.
\end{equation*}
Thus, the expectation of the matrix $\A^T\A$ is easily computed as
\begin{eqnarray*}
	\E[\A^T\A] = \begin{bmatrix}nr_0^2&0&0&nr_0\frac{a^2}{4}\\0&\frac{n}{2}(r_0^2+\frac{a^2}{4})&0&0\\0&0&\frac{n}{2}(r_0^2+\frac{a^2}{4})&0\\nr_0\frac{a^2}{4}&0&0&n(\frac{a^2}{16}+\frac{r_0^2a^2}{3})\end{bmatrix}\;.
\end{eqnarray*}
Let the largest and smallest singular values of $\E[\A^T\A]$ to be $n\sigma_{\max}(r_0,a)$ and $n\sigma_{\min}(r_0,a)$. Using the fact that $\sigma_i(\cdot)$ is a Lipschitz continuous function of
its arguments, together with the Chernoff bound for large deviation of sums of i.i.d. random variables, we get
\begin{eqnarray}
	\prob( \sigma_1(\A\A^T) > 2n\sigma_{\max}(r_0,a)) \leq e^{-Cn} \;,\nonumber\\
	\prob( \sigma_1(\A\A^T) < (1/2)n\sigma_{\max}(r_0,a)) \leq e^{-Cn} \;,\label{eq:2norm2}\\
	\prob( \sigma_4(\A\A^T) < (1/2)n\sigma_{\min}(r_0,a)) \leq e^{-Cn} \;, \label{eq:2norm3}
\end{eqnarray}
for some constant $C$. Hence, with high probability, 
$\kappa(\bD)\leq \frac{4\sigma_{\max}(r_0, a)}{\sigma_{\min}(r_0,a)} = f_{\kappa}(r_0,a)$.

Now to bound $\mu$, note that with probability $1$ the columns of $\A$ are linearly independent.  
Therefore, there exists a matrix $\B\in\R^{r\times r}$ such that 
$\A = \V\B^T$ with $\V^T\V=\eye$.
The SVD of $\bD$ then reads $\bD=\U\Sigma \U^T$ with $\Sigma=\Q^T\B^T\S\B\Q$ 
and $\U = \V\Q$ for some orthogonal matrix $\Q$.
To show incoherence property {\bf A0}, we need to show that, for all $i\in[n]$, 
\begin{eqnarray*}
	\|\V_i\|^2\leq \frac{4\mu}{n}\;.
\end{eqnarray*}
Since $\V_i=\B^{-1}\A_i$, 
we have $\|\V_i\|^2 \leq \sigma_4(\B)^{-2}\|\A_i\|^2\leq \sigma_4(\A)^{-2}\|\A_i\|^2$. 
Combined with $\|\A_i\|^2=r_0^2+(r_0+\rho_i)^2+(2r_0\rho_i+\rho_i^2)^2 \leq r_0^2+(r_0+a)^2+(2r_0a+a^2)^2$ and \eqref{eq:2norm3}, we have 
\begin{eqnarray}
	\|\U_i\|^2 \leq\frac{f_{\mu}(r_0,a)}{n} \;, \label{eq:incoprf1}
\end{eqnarray}
with high probability, where $f_{\mu}(r_0,a) = 2(r_0^2+(r_0+a)^2+(2r_0a+a^2)^2)$.

To show incoherence property {\bf A1}, 
we use $|\bD_{ij}|\leq (2r_0+a)^2$ and $\sigma_1(\bD)\geq \frac{1}{4} n\,\sigma_{\min}(r_0,a)\min\left(2, \frac{\sqrt{1+r_0^2}-r_0}{r_0}\right)$ from \eqref{eq:2norm2}.
Then, 
\begin{eqnarray}
	\frac{|\bD_{ij}|}{\sigma_1(\bD)} \leq \frac{g(r_0,a)}{n} \;, \label{eq:incoprf2}
\end{eqnarray}
with high probability, where $g(r_0, a) = \max\left(2, \frac{4r_0}{\sqrt{1+r_0^2}-r_0}\right)(2r_0+a)^2 / \sigma_{\min}(r_0,a)$. Combining \eqref{eq:incoprf1} and \eqref{eq:incoprf2}, we see that 
the incoherence property is satisfied, with high probability.

Further, \eqref{eq:thm1proof3} holds, with high probability, if 
the right-hand side of \eqref{eq:mainthm1} is less than $C_3\frac{\sqrt{1+r_0^2}+r_0}{r_0}\sigma_{\max}(r_0,a)$, 
since $\sigma_4(\bD)\leq \frac{1}{2}n\frac{\sqrt{1+r_0^2}+r_0}{r_0}\sigma_{\max}(r_0,a)$. 
This finishes the proof of Theorem \ref{thm:main1}.

\myqed
\end{section}

%%%%%%%%%%%%%%%%% Experimental Results %%%%%%%%%%%%%%
\vspace{-0.4cm}
\begin{section}{Simulation Results}
\label{sec:experimental_results}
In order to evaluate the performance of the calibration method, three sets of experiments are done. First, the distance matrix is assumed noiseless and the value of the  $d_0$ is set to zero. The position estimation error is derived for different values of $n$ and the ring width $a$. The value of $r_0$ is set to $10$ cm, on average 5 percent of entries are missing randomly, and $\delta$ in Theorem \ref{thm:main1} is assumed to be 1.  For each value of $a$ and $n$, the experiment is repeated 10 times, and the average is taken. The results are reported in Fig.~\ref{fig:a_diff}. As expected from Corollary~\ref{thm:main2}, the greneral trend in all curves is that the error decreases as $n$ grows. Moreover, the larger $a$ is, the bigger is the reconstruction error, which is also coherent with the results of Corollary~\ref{thm:main2}.
\begin{figure}[tb]
\centering
\scalebox{.95}{
\psfrag{a}{\small$a = 2$ \hfill mm}
\psfrag{b}{\small$a = 5$ \hfill mm}
\psfrag{c}{\small$a = 10$ \hfill mm}
\psfrag{d}{\small$a = 20$ \hfill mm}
\psfrag{v}{\small$10^{-9}$}
\psfrag{r}{\small$10^{-8}$}
\psfrag{q}{\small$10^{-7}$}
\psfrag{s}{\small$10^{-6}$}
\psfrag{m}{\small 200}
\psfrag{n}{\small 700}
\psfrag{o}{\small 1200}
\psfrag{l}{\small 1700}
\psfrag{x}{\small $d(\X,\hX)$}
\psfrag{f}{$n$}
\includegraphics[width = 0.65 \linewidth]{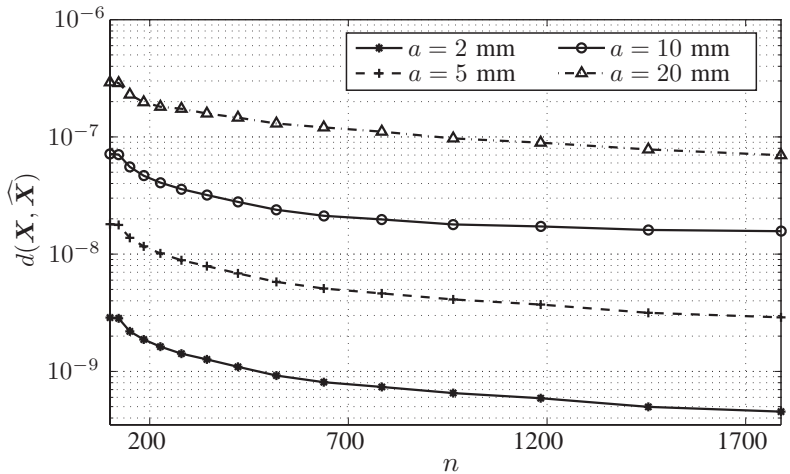}}
\caption{Error in position estimation in noiseless case for different values of $a$. As $n$ increases, the reconstruction error tends to zero. The estimation error increases for larger values of $a$, which confirms the results of Lemma \ref{lem:2norm}.}
\label{fig:a_diff}
\end{figure}

To examine the stability of the estimation algorithm under noise, we set the values of $a$ to $1$ cm, $\delta$ to $1$, $r_0$ to $10$ cm, $t_0$ to zero, and the percentage of random missing entries to 5. We added to each entry of the distance matrix $\D$ a centred white Gaussian noise of different standard deviations. For each $n$ and standard deviation of noise, the experiments are repeated 10 times and the average is taken. The results are depicted in Fig.~\ref{fig:noise_diff}\ \footnote{There has been a slight mislabeling in the earlier version of this paper in \cite{par11} which is corrected in this paper.} . As the variance of the noise increases, the position estimation error grows, but in general the error decreases for larger $n$. 

\begin{figure}[tb]
\centering
\scalebox{0.95}{
\psfrag{a}{\small$\sigma = 0.6$ mm}
\psfrag{b}{\small$\sigma = 3\;\,\,\,$ mm}
\psfrag{c}{\small$\sigma = 6\,\,\,$ mm}
\psfrag{d}{\small$\sigma = 10$ mm}
\psfrag{p}{\small$10^{-6}$}
\psfrag{q}{\small$10^{-5}$}
\psfrag{r}{\small$10^{-4}$}
\psfrag{s}{\small$10^{-3}$}
\psfrag{h}{\small 200}
\psfrag{g}{\small 700}
\psfrag{m}{\small 1200}
\psfrag{w}{\small 1700}
\psfrag{x}{\small $d(\X,\hX)$}
\psfrag{f}{$n$}
\includegraphics[width = 0.65 \linewidth]{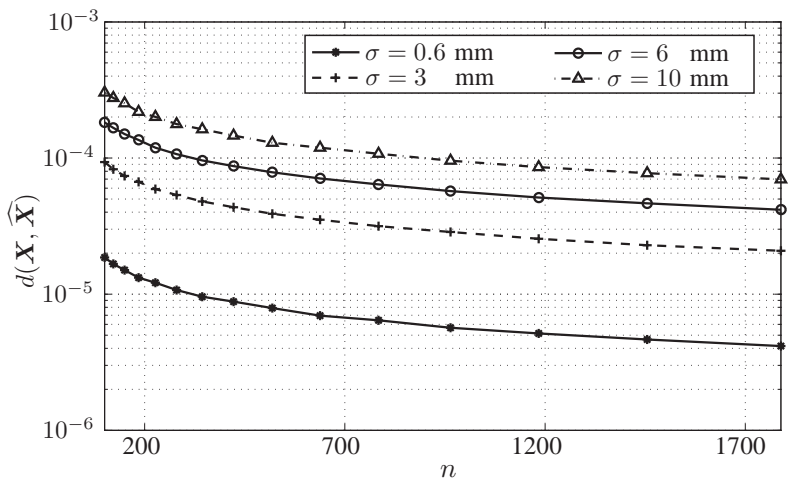}}
\caption{Error in position estimation for the case with centered white Gaussian noise of different standard deviations, $\sigma$.}
\label{fig:noise_diff}
\end{figure}

As we discussed in Section \ref{sec:relatedWork}, one might treat the calibration problem as a special case of the sensor localization problem. However, there is a duality between the calibration problem and the traditional sensor localization problems. We showed in Section \ref{sec:mod_def} that in the calibration problem the local distance/connectivity information is not available whereas most of the state-of-the-art algorithms for sensor localization are based on the local information. In order to compare the performance of these methods with the proposed methods, a set of simulations are performed. We compared the localization results of our method to the ones of \textsc{Mds-Map} \cite{SRZ03}, \textsc{Sdp}-based \cite{SDP06} and also \textsc{Svd-Reconstruct} \cite{dri06}. The position reconstruction error (defined in  \eqref{distance}) versus the number of sensors, $n$ for the methods is reported in Fig.~\ref{fig:comparison} in a log-log scale. 

\begin{figure}[tb]
\centering
\scalebox{0.95}{
\providecommand\matlabtextA{\color[rgb]{0.000,0.000,0.000}\fontsize{10}{10}\selectfont\strut}%
\psfrag{n}[cl][cl]{\footnotesize\sc Svd-Reconstruct}%
\psfrag{m}[cl][cl]{\footnotesize\sc Mds-Map}%
\psfrag{l}[cl][cl]{\footnotesize\sc Sdp}%
\psfrag{k}[cl][cl]{\footnotesize\sc Our Method}%
\psfrag{o}[bc][bc]{\matlabtextA $d(\X,\hX$)}%
\psfrag{p}[tc][tc]{\matlabtextA $n$}%
\def\matlabfragNegXTick{\mathord{\makebox[0pt][r]{$-$}}}
\psfrag{a}[Bc][Bc]{\matlabtextA $25$}%
\psfrag{b}[Bc][Bc]{\matlabtextA $50$}%
\psfrag{c}[Bc][Bc]{\matlabtextA $100$}%
\psfrag{d}[Bc][Bc]{\matlabtextA $200$}%
\psfrag{e}[Bc][Bc]{\matlabtextA $400$}%
\psfrag{f}[Bc][Bc]{\matlabtextA $800$}%
\psfrag{g}[Bc][Bc]{\matlabtextA $1600$}%
\psfrag{h}[rc][rc]{\matlabtextA $10^{-5}$}%
\psfrag{i}[rc][rc]{\matlabtextA $10^{-4}$}%
\psfrag{j}[rc][rc]{\matlabtextA $10^{-3}$}%
\includegraphics[width = 0.65 \linewidth]{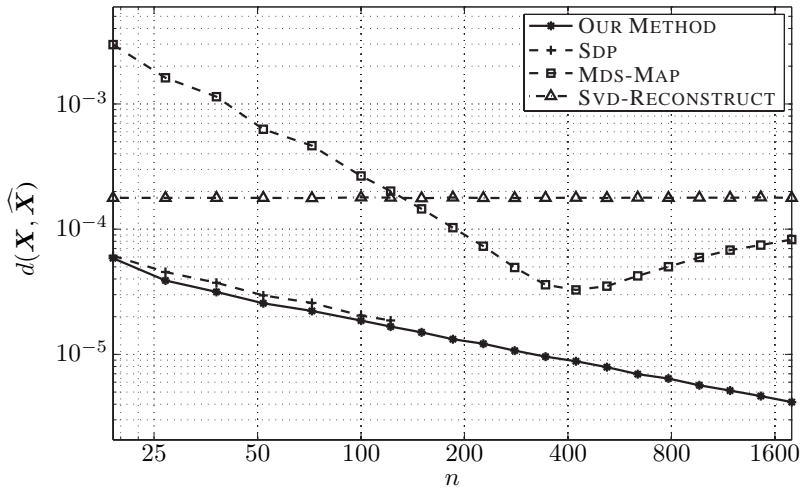}}
\caption{Error in position estimation versus the number of sensors for different methods. }
\label{fig:comparison}
\end{figure}

For the simulations, we set the values of $a$ to 1 cm, $\delta$ to 1, $r_0$ to 10 cm, $t_0$ to zero, and the percentage of the random missing entries to 5. The distance measurements were corrupted with a white Gaussian noise of standard deviation $0.6$ mm. For each method and each $n$, the experiment is performed 10 times for different positions and different noises, and the average error is taken. For the \textsc{Sdp}-based method, we have used the algorithm presented in \cite{SDP06} and the code published by the same authors. For \textsc{Mds-Map}, we have estimated the shortest paths using Johnson's algorithm \cite{john77}. Finally for \textsc{Svd-Reconstruct}, we used the algorithm in \cite{dri06}. In order to adapt the measurements with the assumptions of the method, we assumed that $p_{ij} = 1-0.05 = 0.95$ for the measured points (note that $0.05$ is on average the probability of having a random missing entry) and $\gamma_{ij} = 0$. 

As the results in Fig.~\ref{fig:comparison} suggest, \textsc{Mds-Map} and \textsc{Svd-Reconstruct} methods perform very bad compared to the other two methods. The poor performance of \textsc{Mds-Map} is for the fact that it highly relies on the presence of local distance information, whereas in our case, these measurements are in fact missing. This method is based on estimating the missing distance measurements with the shortest path between the two sensors. However, one can easily see that given faraway sensor distances, the shortest path is a very coarse approximation of the distance between close-by sensors. also note that as the simulation results show, there is no guarantee that the estimation error will decrease as $n$ grows.

For \textsc{Svd-Reconstruct}, the unrealistic assumption that all the sensors have a non-zero probability of being connected causes the bad results of the method. In our case, the probability that the close-by sensors are connected is zero because of the structured missing entries. In fact, since $p_{ij}$ is high, one could see this method as simply applying the classical MDS on the incomplete distance matrix. The surprising observation about the performance of this method is that the estimation error does not change much with $n$. 

In contrast to the two aforementioned algorithms, the \textsc{Sdp}-based method performs very well for estimating the sensor positions and the reconstruction error is very close to the one of the proposed method. The reason is the fact that this method does not directly rely on the local distance information. In fact, the distance measurements are fed to the algorithm as the constraints of a convex optimization problem. However, as the number of sensors goes large, the number of measurements also grows and so does the number of constraints for the semi-definite program. This causes the method not to work for $n$ larger than 150 in our case. The same limitation is also reported by the authors of the method. 

In summary, taking the computational cost and reconstruction accuracy of the algorithms into account, the proposed method performs significantly better.

Moreover, to show the importance of calibration in an ultrasound scanning device, a simple simulation is also performed. If the ToF measurements correspond to the exact positions of sensors without time delay $t_0$, reconstruction of water will lead to a homogeneous region with values equal to the water sound speed, whereas wrong assumption on the sensor positions and $t_0$ causes the inverse method to give incorrect values as the sound speed to compensate the effect of position mismatch. 

In a simple experiment, we simulated the reconstruction of water sound speed ($c_0 = 1500$) using the ToF measurements. In the simulation, 200 sensors are distributed around a circle with radius $r_0 = 10$ cm, and they deviate at most $5$ mm from the circumference and the ToF measurements are added by $t_0 = 10 \mu s$.  The incomplete distance matrix is shown in Fig.~\ref{fig:tof_incomplete}. 

In order to complete the distance matrix and find the time delay at the same time, we used Algorithm~\ref{alg:find_t_0}. We forced the rank of $\bD$ to 4. The value for $t_0$ is found as $4 \mu s$ which is exactly as set in the simulation.  The output of \opt algorithm is the completed $\bD$ matrix which is shown in Fig.~\ref{fig:tof_complete}. 

\begin{figure}[tb]
\centering
\subfigure[Incomplete distance squared matrix]{
\includegraphics[height = 5cm]{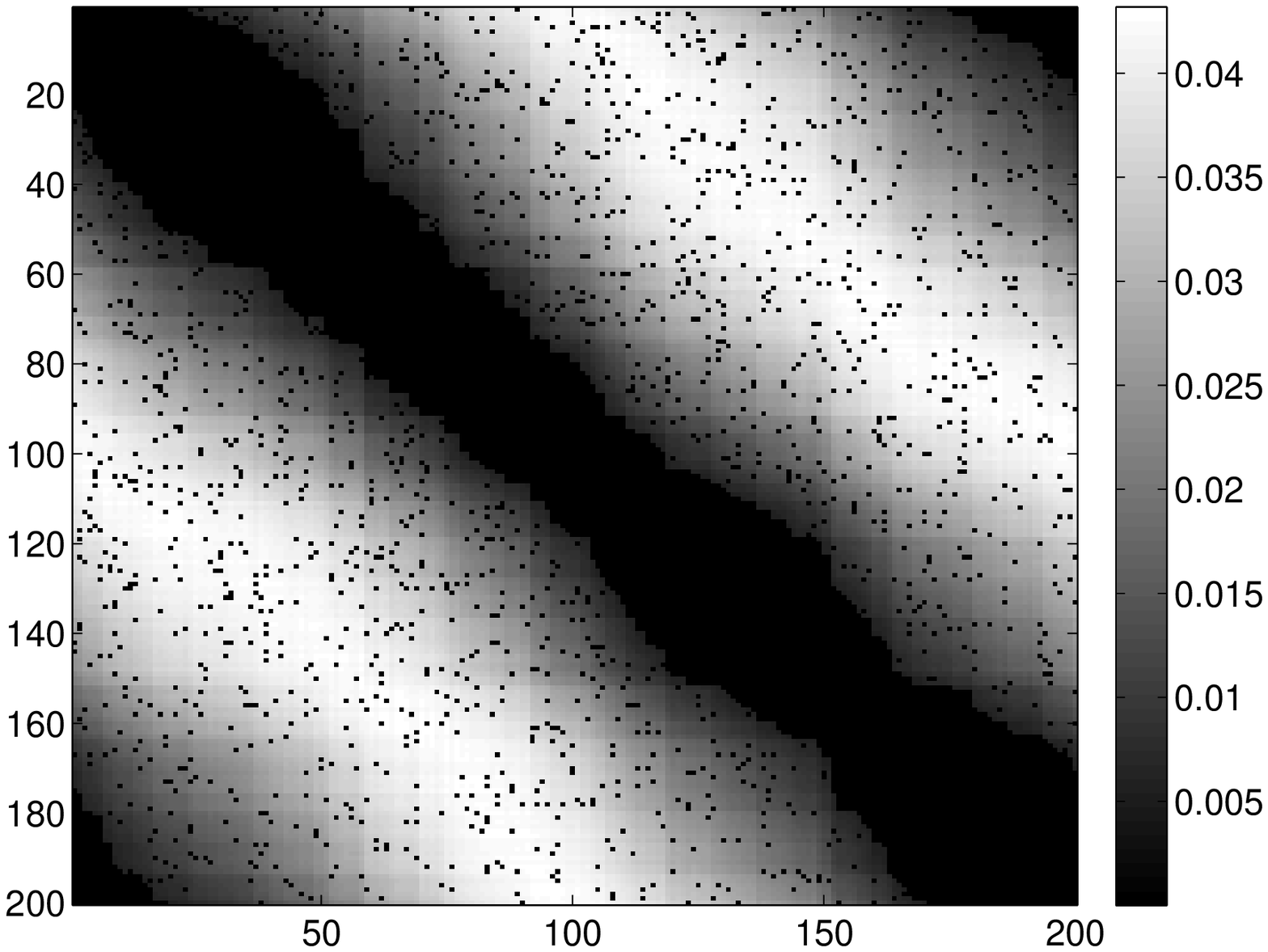}
\label{fig:tof_incomplete}
}
\subfigure[Completed distance squared matrix]{
\includegraphics[height=5cm]{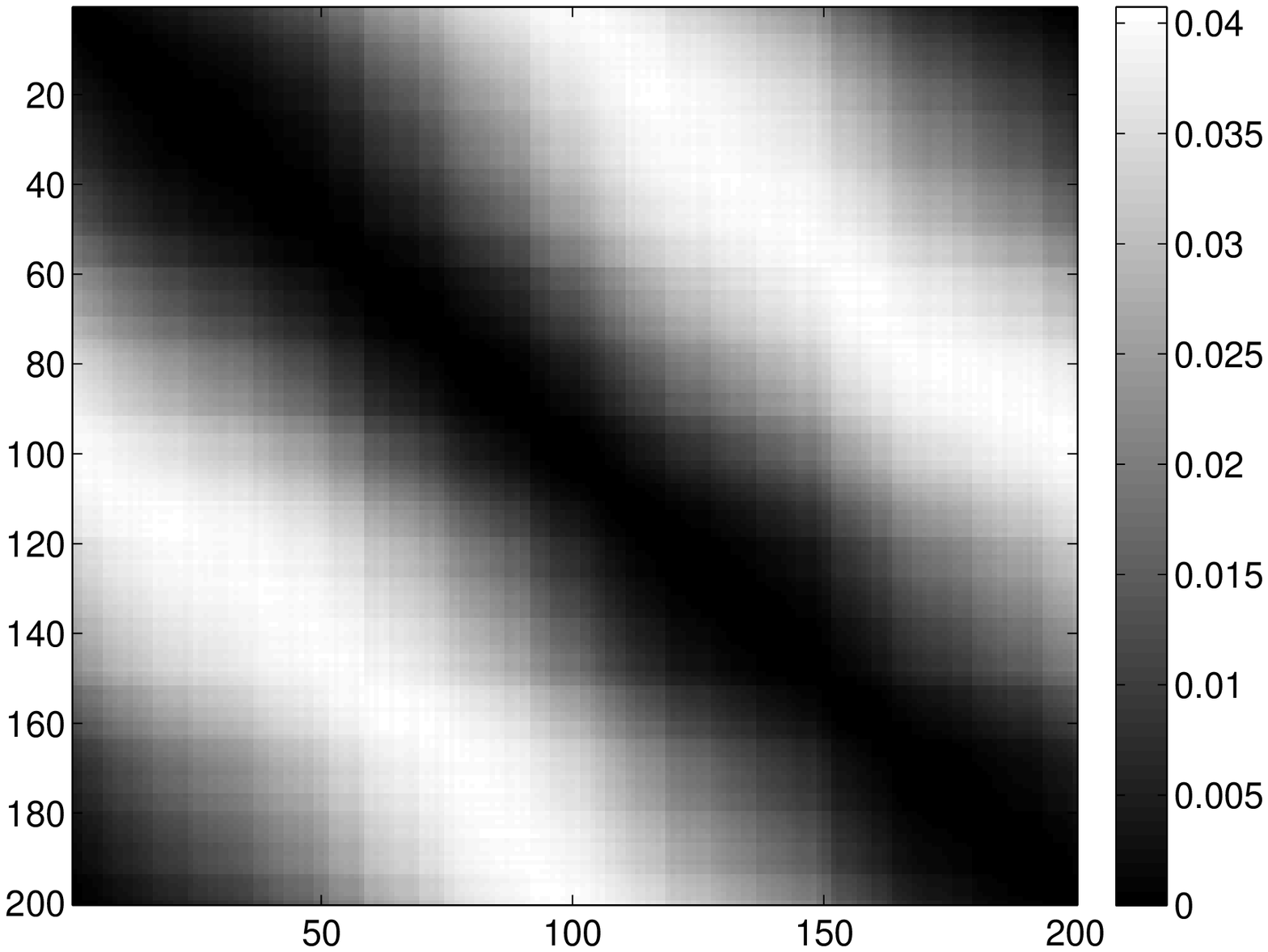}
\label{fig:tof_complete}
}
\label{fig:tof_matrices}
\caption{Input and output of \opt algorithm. \subref{fig:tof_incomplete} The incomplete distance squared matrix $\bD$, with 5 percent of entries randomly missing, $t_0 = 10\mu s$ and $\delta_n = 3 cm$. \subref{fig:tof_complete} The completed matrix with estimated $t_0 = 10 \mu s$. The modified \opt algorithm in this case can find the time mismatch correctly.}
\end{figure}

Using the completed distance matrix and the MDS method, the positions are reconstructed and fed to an inverse tomography algorithm to reconstruct water sound speed. The results of the reconstruction are shown in Fig.~\ref{fig:inverse_outputs}. In the figure, the results for four reconstructions are presented. In Fig.~\ref{fig:uncalibrated_water}, the ToF matrix is not complete, it contains the time delay $t_0$, and the positions are not calibrated. The dark gray ring is caused by the non-zero time delay in the ToF measurements. In Fig.~\ref{fig:t0_0_def_pos}, the time mismatch is resolved using the proposed algorithm, but the sensor positions are not calibrated and the ToF matrix is still not complete. This figure shows clearly that finding the unknown time delay improves significantly the reconstruction image. Figure \ref{fig:complete_t0_0_def_pos}, shows the reconstructed medium when the ToF matrix is completed and time mismatch is removed, but the sensor positions are not yet calibrated. From this figure, it is confirmed that accurate time-of-flights are necessary but not sufficient to have a good reconstruction of the inclosed object. Finally, Fig.~\ref{fig:estimated_t0_0_good_pos} shows the reconstruction when the positions are also calibrated. Notice the change in the dynamic range for the last case.

\begin{figure}[tb]
\centering
%\ContinuedFloat
%\subfigure[When no calibration is done. The time mismatch is $t_0 = 4e-6$ and the positions are assumed to be on a circle.]{
\subfigure[]{\psfrag{s10}[][]{\color[rgb]{0,0,0}\setlength{\tabcolsep}{0pt}\begin{tabular}{c} \end{tabular}}%
\psfrag{s11}[][]{\color[rgb]{0,0,0}\setlength{\tabcolsep}{0pt}\begin{tabular}{c} \end{tabular}}%
%
% xticklabels:
\psfrag{x01}[t][t]{-0.1}%
\psfrag{x02}[t][t]{-0.05}%
\psfrag{x03}[t][t]{0}%
\psfrag{x04}[t][t]{0.05}%
\psfrag{x05}[t][t]{0.1}%
%
% yticklabels:
\psfrag{v01}[l][l]{1200}%
\psfrag{v02}[l][l]{1500}%
\psfrag{v03}[l][l]{1800}%
\psfrag{v04}[r][r]{-0.1}%
\psfrag{v05}[r][r]{-0.05}%
\psfrag{v06}[r][r]{0}%
\psfrag{v07}[r][r]{0.05}%
\psfrag{v08}[r][r]{0.1}%
%
% Figure:
\resizebox{0.45\linewidth}{!}{\includegraphics{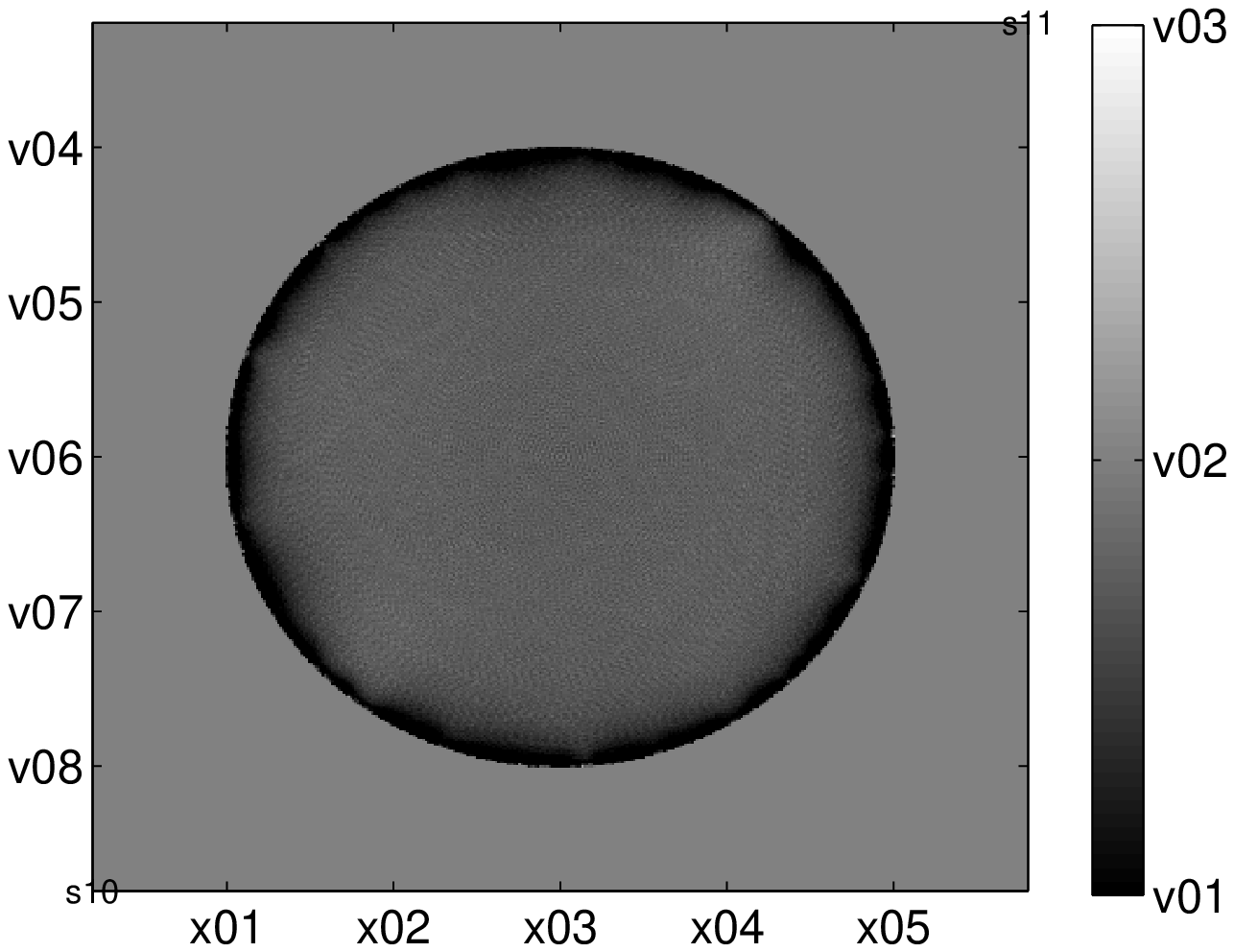}}%
\label{fig:uncalibrated_water}
}
%\subfigure[When the time mismatch is found and removed from the measurements, but the ToF matrix is incomplete and the positions are not correct]{
\subfigure[]{
\psfrag{s10}[][]{\color[rgb]{0,0,0}\setlength{\tabcolsep}{0pt}\begin{tabular}{c} \end{tabular}}%
\psfrag{s11}[][]{\color[rgb]{0,0,0}\setlength{\tabcolsep}{0pt}\begin{tabular}{c} \end{tabular}}%
%
% xticklabels:
\psfrag{x01}[t][t]{-0.1}%
\psfrag{x02}[t][t]{-0.05}%
\psfrag{x03}[t][t]{0}%
\psfrag{x04}[t][t]{0.05}%
\psfrag{x05}[t][t]{0.1}%
%
% yticklabels:
\psfrag{v01}[l][l]{1200}%
\psfrag{v02}[l][l]{1500}%
\psfrag{v03}[l][l]{1800}%
\psfrag{v04}[r][r]{-0.1}%
\psfrag{v05}[r][r]{-0.05}%
\psfrag{v06}[r][r]{0}%
\psfrag{v07}[r][r]{0.05}%
\psfrag{v08}[r][r]{0.1}%
%
% Figure:
\resizebox{0.45\linewidth}{!}{\includegraphics{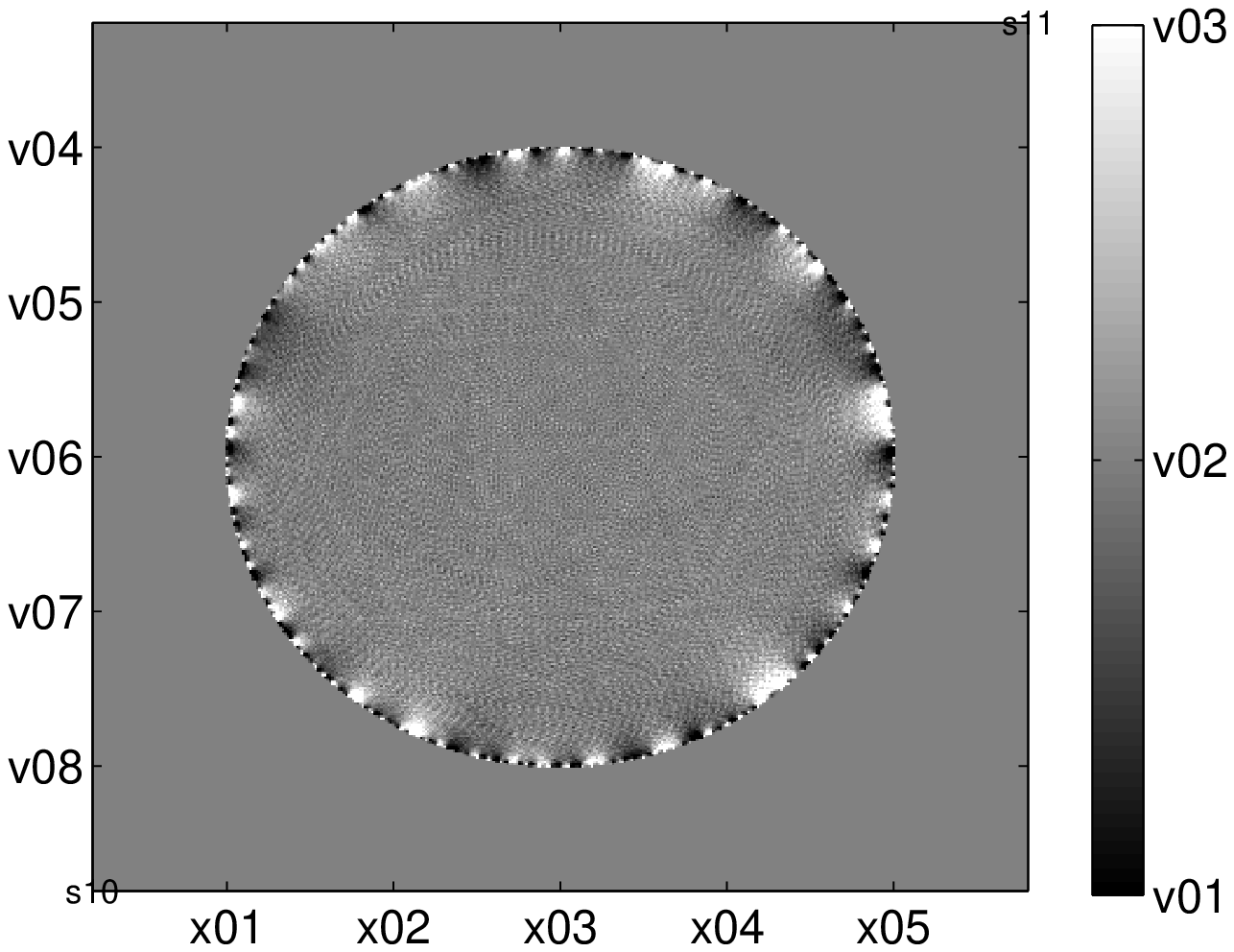}}%
\label{fig:t0_0_def_pos}
}\\
%\subfigure[When the time mismatch is found and removed from the measurements, the ToF matrix is estimated but the positions are not correct]{
\subfigure[]{
\psfrag{s10}[][]{\color[rgb]{0,0,0}\setlength{\tabcolsep}{0pt}\begin{tabular}{c} \end{tabular}}%
\psfrag{s11}[][]{\color[rgb]{0,0,0}\setlength{\tabcolsep}{0pt}\begin{tabular}{c} \end{tabular}}%
%
% xticklabels:
\psfrag{x01}[t][t]{-0.1}%
\psfrag{x02}[t][t]{-0.05}%
\psfrag{x03}[t][t]{0}%
\psfrag{x04}[t][t]{0.05}%
\psfrag{x05}[t][t]{0.1}%
%
% yticklabels:
\psfrag{v01}[l][l]{1200}%
\psfrag{v02}[l][l]{1500}%
\psfrag{v03}[l][l]{1800}%
\psfrag{v04}[r][r]{-0.1}%
\psfrag{v05}[r][r]{-0.05}%
\psfrag{v06}[r][r]{0}%
\psfrag{v07}[r][r]{0.05}%
\psfrag{v08}[r][r]{0.1}%
%
% Figure:
\resizebox{0.45\linewidth}{!}{\includegraphics{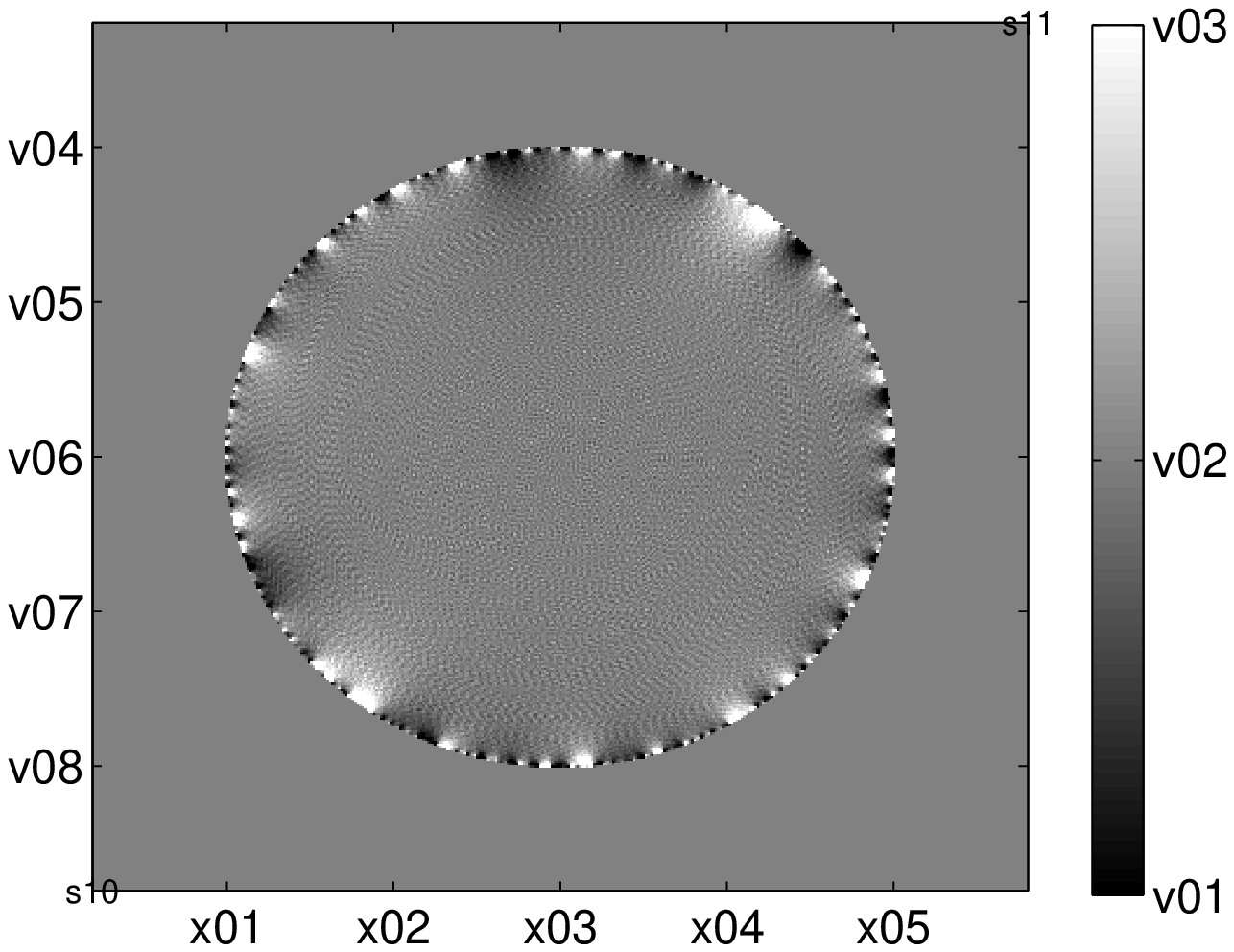}}
\label{fig:complete_t0_0_def_pos}
}
%\subfigure[When the time mismatch is found and removed from the measurements, the ToF matrix is estimated and the positions estimated]{
\subfigure[]{\psfrag{s10}[][]{\color[rgb]{0,0,0}\setlength{\tabcolsep}{0pt}\begin{tabular}{c} \end{tabular}}%
\psfrag{s11}[][]{\color[rgb]{0,0,0}\setlength{\tabcolsep}{0pt}\begin{tabular}{c} \end{tabular}}%
%
% xticklabels:
\psfrag{x01}[t][t]{-0.1}%
\psfrag{x02}[t][t]{-0.05}%
\psfrag{x03}[t][t]{0}%
\psfrag{x04}[t][t]{0.05}%
\psfrag{x05}[t][t]{0.1}%
%
% yticklabels:
\psfrag{v01}[l][l]{1490}%
\psfrag{v02}[l][l]{1500}%
\psfrag{v03}[l][l]{1510}%
\psfrag{v04}[r][r]{-0.1}%
\psfrag{v05}[r][r]{-0.05}%
\psfrag{v06}[r][r]{0}%
\psfrag{v07}[r][r]{0.05}%
\psfrag{v08}[r][r]{0.1}%
%
% Figure:
\resizebox{0.45\linewidth}{!}{\includegraphics{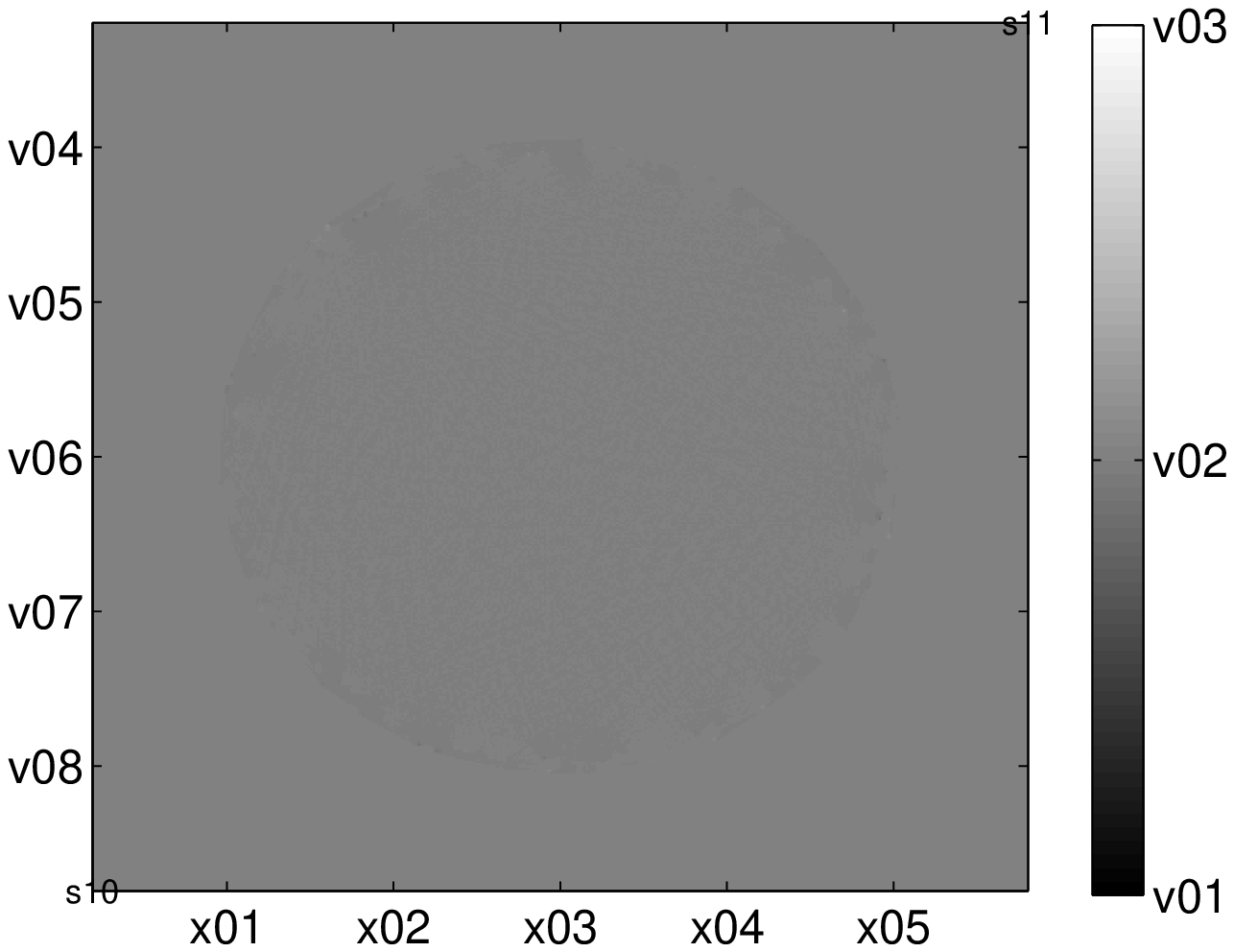}}%
\label{fig:estimated_t0_0_good_pos}
}
\caption{Results of the inversion procedure for finding the sound speed inside the ring with only water inside. \subref{fig:uncalibrated_water} Reconstruction of homogeneous water when no calibration in performed. \subref{fig:t0_0_def_pos} Same after $t_0$ is removed from the ToF matrix, but the matrix is still incomplete and the positions are not calibrated. \subref{fig:complete_t0_0_def_pos} Reconstruction when the matrix is also completed, but the positions are not yet calibrated. \subref{fig:estimated_t0_0_good_pos} Reconstruction with completed ToF matrix and calibrated positions.}
\label{fig:inverse_outputs}
\end{figure}
\end{section}

\section{Conclusion and Future Work}
In this work we introduced a theoretical framework for calibration in circular ultrasound tomography devices. We proposed a novel calibration algorithm for which we provided theoretical bounds on the performance. We also tested our method through exhaustive simulations to demonstrate its functionality in practice. We compared the algorithm with some state-of-the-art centralized sensor localization methods and showed that our method outperforms those in estimating the correct sensor positions. 

Even though we introduced a recursive algorithm for finding the time-delay, we were not able to provide theoretical guarantees on its convergence. We mainly observed its convergence through simulations. This is still an interesting theoretical challenge and requires further work. We also believe that our approach can potentially be deployed beyond circular to other popular topologies with simple geometry. 
%\section{Reproducible Research}
\bibliographystyle{IEEEtran} 
\bibliography{references}
\appendix
\vspace{-.4cm}
\subsection{Proof of Lemma~\ref{lem:rank3}}\label{proof:rank3}
The proof for the general case where the sensors are not on a circle is provided in \cite{dri06}. In the circular case however, we have $\bD_{i,j} = \norm{\x_i}^2 + \norm{\x_j}^2 - 2 \x_i^T\x_j = 2r^2 - 2\x_i^T\x_j$, where $r$ is the circle radius. Thus, the squared distance matrix is decomposable to 
\begin{equation*}
\bD = \V \bm{\Sigma}\V^T\,,
\end{equation*}
where 
\begin{equation*}
\V= \begin{bmatrix}
r & x_{1,1} & x_{1,2}\\
\vdots & \vdots & \vdots\\
r & x_{n,1} & x_{n,2}
\end{bmatrix} \,, \quad \bm{\Sigma}  = \begin{bmatrix}
2 & 0 & 0\\
0 & -2 & 0\\
0 & 0 & -2
\end{bmatrix}\,.
\end{equation*}
This finishes the proof. \myqed
%%%%%%%%%%%%%%%%%%%%%%%%%%%%%%%%%%%%%%%%%%%%%%%%%%%%%%%%%%%%%%%%%%%%%%%%%%%%%
\vspace{-.4cm}
\subsection{Proof of Corollary \ref{thm:main2}}\label{proof:thm2}
Note that in general $(\L \X\X^T\L-\L\hX\hX^T\L)$ has rank at most $2d$ where $d$ is the dimension of the space in which sensors are placed (in our case $d=2$). Therefore, 
$$ \norm{\L \X\X^T\L-\L\hX\hX^T\L }_F \leq \sqrt{2d}\norm{\L \X\X^T\L-\L\hX\hX^T\L }_2,$$where we used the fact that for any matrix $A$ of rank $r$ we have $\norm{A}_F\leq \sqrt{r}\norm{A}_2$. Furthermore, the spectral norm can be bounded in terms of $\bD$ and $\widehat{\bD}$ as follows. 
\begin{eqnarray}
\norm{\L \X\X^T\L-\L\hX\hX^T\L }_2 &\stackrel{(a)}{\leq}& \norm{\L \X\X^T\L-\frac{1}{2}\L\widehat{\bD}\L}_2+\norm{\frac{1}{2}\L\widehat{\bD}\L-\hX\hX^T}_2 \nonumber\\
 &\stackrel{(b)}{\leq}& \frac{1}{2} \norm{\L(\bD-\widehat{\bD})\L}_2 +  \frac{1}{2} \norm{\L(-\bD+\widehat{\bD})\L}_2, \label{spectral}
\end{eqnarray}
where in $(a)$, we used the triangle inequality and \eqref{LMDS}, namely, $\L\hX=\hX$. In $(b)$, we used \eqref{LMDS2} and the fact that for any matrix $A$ of rank $d$, $\norm{\frac{1}{2}\L\widehat{\bD}\L -\hX\hX^T}_2 \leq \norm{\frac{1}{2}\L\widehat{\bD}\L-A}_2$. In particular, by setting $A=\frac{1}{2}\L\bD\L$ the second term in \eqref{spectral} follows. 
Since $\L$ is a projection matrix we have $\norm{\L}_2=1$. Hence, from \eqref{spectral} we can conclude that  $$\norm{\L \X\X^T\L-\L\hX\hX^T\L }_2\leq \norm{\widehat{\bD} -\bD}_2.$$ This immediately leads to Corollary~\ref{thm:main2}. \myqed
%Since $\L$ is a projection, we have $\|\L\X\X^T\L-\L\Xh\Xh^T\L\|_F=(1/2)\|\L\bD\L-\L\Dh\L\|_F\leq(1/2)\|\bD-\Dh\|_F$, 
%it follows that $d_1(\Xh,\X)\leq (1/2n)\|\bD-\Dh\|_F$. 
%Therefore, the claim follows directly from Theorem \ref{thm:main1}.

%%%%%%%%%%%%%%%%%%%%%%%%%%%%%%%%%%%%%%%%%%%%%%%%%%%%%%%%%%%%%%%%%%%%%%%%%%%%%%%%%%
\vspace{-.4cm}
\subsection{Proof of Lemma~\ref{lem:2norm}}\label{proof:2norm}
Note that by the definition of $\bD^s$, we have 
$|\cP_E(\bD^s)_{i,j}| \leq \delta_n^2$ for all $i$ and $j$.
Define $\A$ as 
\begin{eqnarray*}
 	\A_{i,j} = 
	\begin{cases}
		\; 1 & \text{if } (i,j) \in E\cap S \;,\\
		\; 0 & \text{otherwise}\;.
	\end{cases}
\end{eqnarray*}
We start from a simple realtionship 
between an elementwise bounded matrix and its operator norm. 
\begin{eqnarray}
	\|\cP_E(\bD^s)\|_2 &\leq& \delta_n^2 \max_{\|x\|=\|y\|=1} \sum_{i,j} |x_i|\, |y_j|\, \A_{i,j} \label{eq:lem_Dbs2} =  \delta_n^2 \|\A\|_2 \; \nonumber.
\end{eqnarray}
The inequlity in (\ref{eq:lem_Dbs2}) follows from the fact 
that $\cP_E(\bD^s)$ is elementwise bounded by $\delta_n$. 
We can further bound the operator norm $\|\A\|_2$, 
by applying the celebrated Gershgorin circle theorem to a symmetrized version of $\A$. 
Define a symmetric matix $\bA$ as 
\begin{eqnarray*}
 	\bA_{i,j} = 
	\begin{cases}
		\; 1 & \text{if } (i,j) \in E\cap S \text{ or } (j,i) \in E\cap S \;,\\
		\; 0 & \text{otherwise}\;.
	\end{cases}
\end{eqnarray*}
Since $0 \leq \A_{i,j} \leq \bA_{i,j}$ for all $i$ and $j$, 
we have $\|\A\|_2 \leq \|\bA\|_2$.
Applying the Gershgorin circle theorem we get
\begin{eqnarray*}
	\|\bA\|_2 \leq \max_{i\in[n]}\sum_{j\in[n]} |\bA_{i,j}| \;. \label{eq:lem_Dbs5}
\end{eqnarray*}
Define random variables $\{Y_1,\ldots,Y_n\}$, where $Y_i$ is the number of 
non-zero entries in the $i$th row of $\bA$.
Then, 
\begin{equation*}
	\|\bA\|_2 \leq \max_{i\in[n]} Y_i \;. \label{eq:lem_Dbs6}
\end{equation*}

We need to show that $Y_i$ concentrates around its mean. 
Since $Y_i$'s are binomial random variables, we can apply the Chernoff bound. Recall that $(i,j)\in S$ if $\|\x_i-\x_j\| \leq \delta_n$. By the definition of $E$, 
each sample is sampled with probability $p$. Then the probability that either $(i,j)$ or $(j,i)$ is in $E$ is $2p-p^2$.

\begin{figure}[tb]
\centering
\psfrag{a}{\small $a$}
\psfrag{b}{\small $r_2$}
\psfrag{c}{\small $r_0$}
\psfrag{d}{\small $r$}
\psfrag{e}{\small $r_1$}
\psfrag{f}{\small $A(r)$}
\psfrag{g}{\small $\delta_n$}
\includegraphics[scale = 0.65]{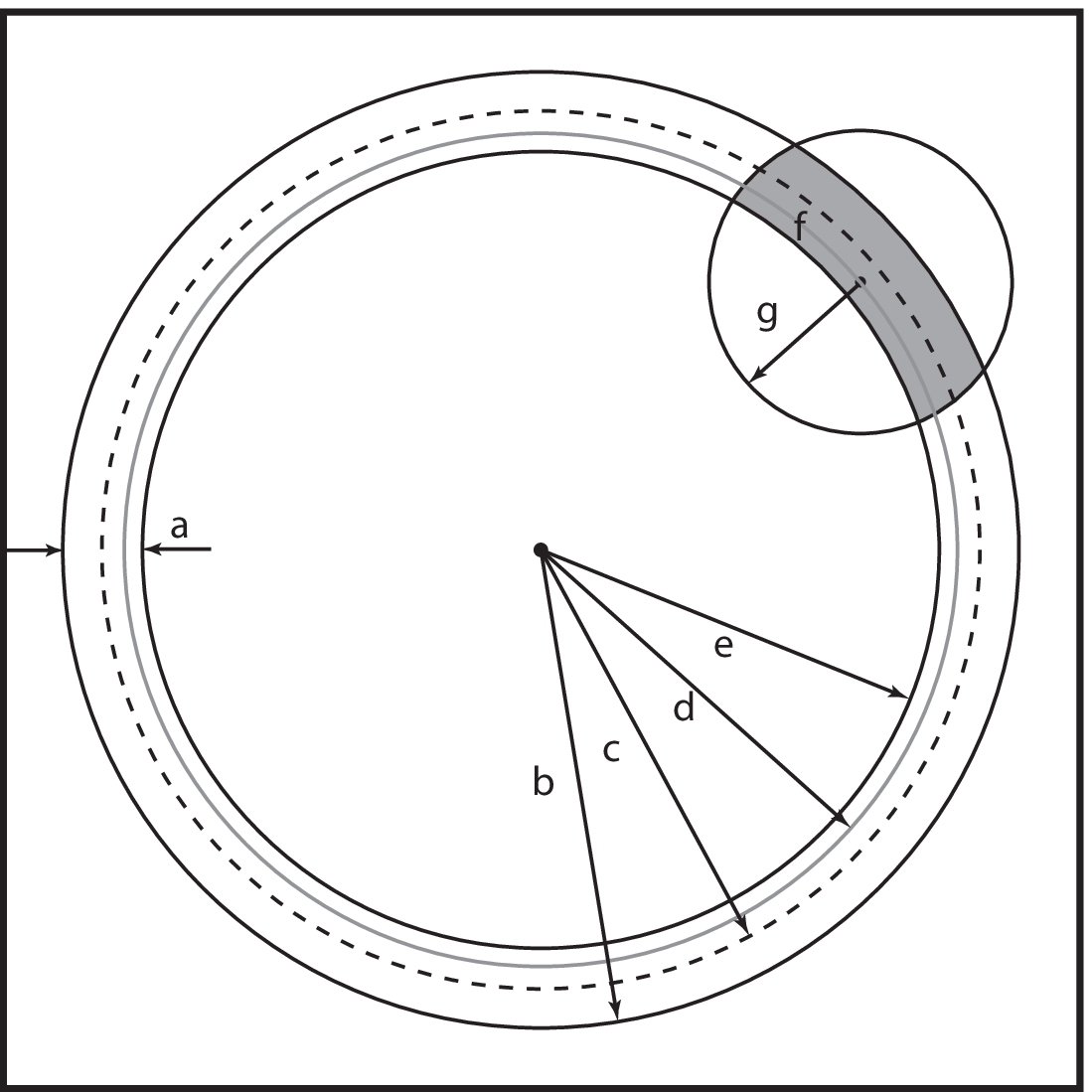}
\caption{The process for bounding the probability of a pair of sensors to fall in $S$. $r_1 = r_0 - a/2$ and $r_2 = r_0 + a/2$.}
\label{fig:integration_circle}
\end{figure}

\begin{figure}[tb]
\centering
\psfrag{a}{\tiny $\frac{\alpha}{2}$}
\psfrag{b}{\footnotesize $\delta_n$}
\psfrag{c}{\footnotesize $r$}
\includegraphics[scale = 0.65]{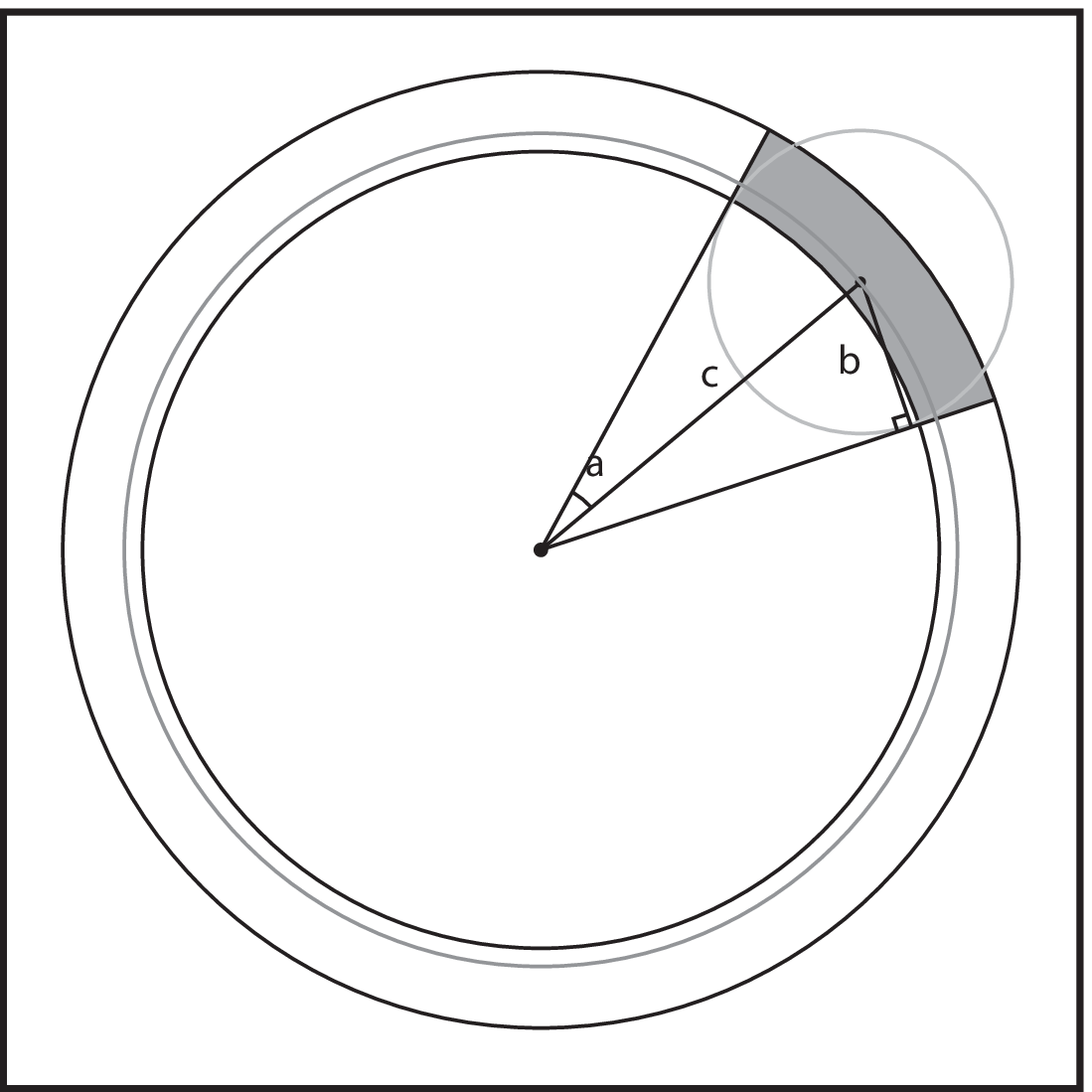}
\caption{Upper bound on $A(r)$. The grey area constructed by the tangents to the $\delta_n$ circle is an upper bound for $A(r)$.}
\label{fig:lower_bound}
\end{figure}

Each entry in the $i$th row of $\bA$ is an independent Bernoulli  random variable with probability of being one equal to $q(2p-p^2)$, where $q$ is the probability that a pair is in $S$. Thus, we have $\E[Y_i] = q(2p-p^2)n$. In order to find the bounds on $\E[Y_i]$, we need to bound $q$. Figure \ref{fig:integration_circle} shows the process for obtaining the bounds on $q$. 
\begin{equation*}
q = \prob\{|\x_i - \x_j| \leq  \delta_n\} = \int_{r_1}^{r_2}\frac{2\pi r}{\pi (r_2^2- r_1^2)}p_2(r)dr\,,
\end{equation*}
where
%\begin{equation*}
$p_2(r) = \frac{A(r)}{\pi(r_2^2-r_1^2)}$.\\
%\end{equation*}
\textbf{Upper Bound on $A(r)$:}\\
Obviously the area $A(r)$ can be bounded by what is shown in Fig.~\ref{fig:lower_bound}. 
Thus, we will have
\begin{equation*}
\sin \frac{\alpha}{2} = \frac{\delta_n}{r}\,.
\end{equation*}
Note that for $0<\alpha<\pi$, $\alpha/\pi\leq \sin \alpha/2\leq \alpha/2$. Hence, $\alpha/\pi\leq \delta_n/r\leq \alpha/2$. So,
\begin{equation*}
A(r) \leq \frac{\alpha}{2\pi}\pi(r_2^2-r_1^2) \leq \frac{\delta_n\pi}{2r}(r_2^2-r_1^2)\,.
\end{equation*}
Thus
\begin{equation*}
p_2(r) \leq \frac{\frac{\delta_n\pi}{2r}(r_2^2-r_1^2)}{\pi(r_2^2-r_1^2)} = \frac{\delta_n}{2r}\,.
\end{equation*}
\begin{equation*}
q \leq \int_{r_1}^{r_2} \frac{2\pi r}{\pi(r_2^2-r_1^2)}\cdot \frac{\delta_n}{2r}dr = \frac{\delta_n}{r_2+r_1} = \frac{\delta_n}{2r_0}\,.
\end{equation*}
\textbf{Lower Bound on $A(r)$:}\\
In order to find the lower bound, we consider the following two different situations: 1) $\delta_n \leq a$ and 2) $\delta_n > a$.

\begin{figure}[tb]
\centering
\subfigure[Lower bound in case 1.]{\psfrag{b}{ \small$a$}\includegraphics[scale = 0.65]{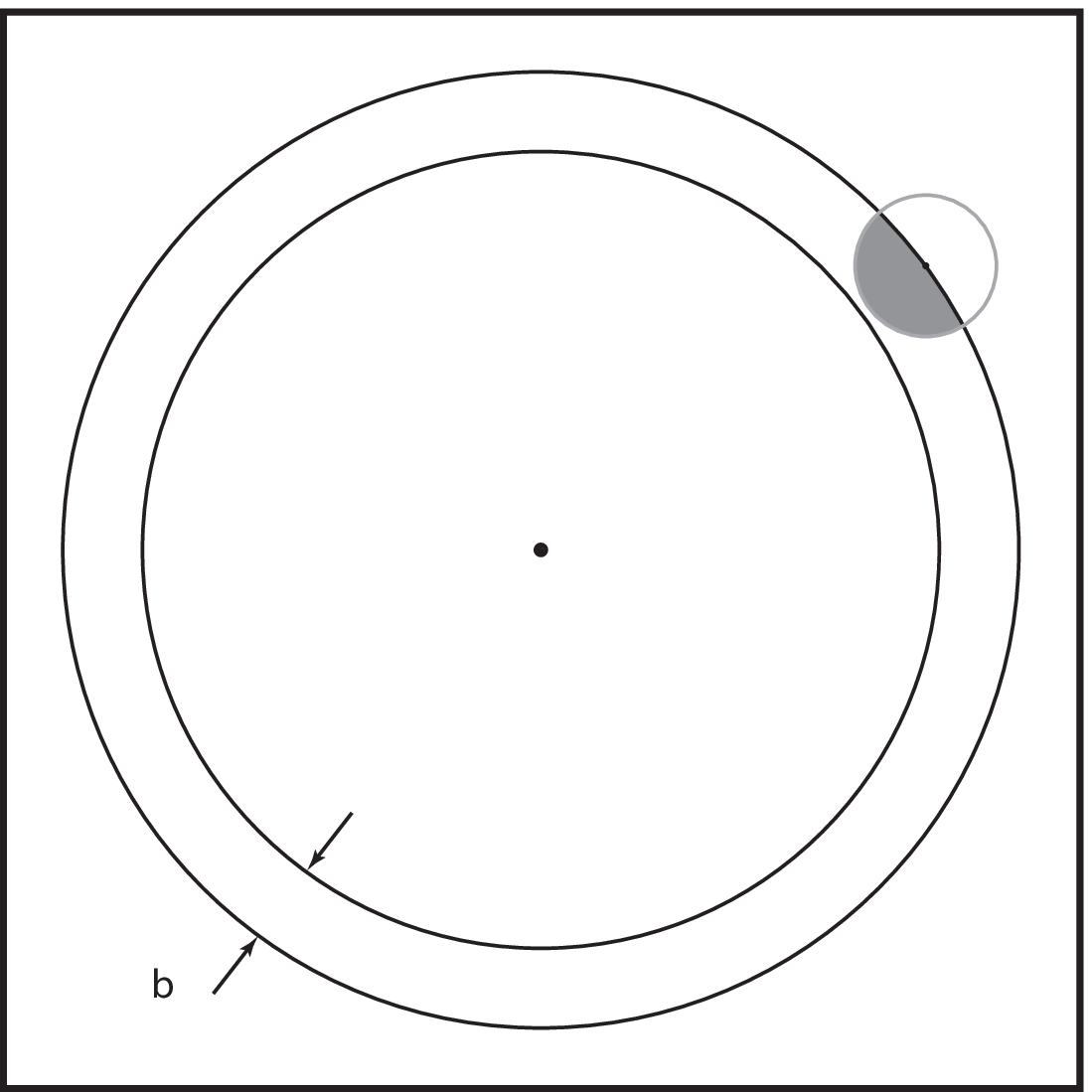}
\label{fig:lower_bound1}
}
\subfigure[Lower bound in case 2.]{\psfrag{a}{\tiny $x_1$}
\psfrag{b}{\tiny  $x_2$}
\psfrag{c}{\tiny  $y_1$}
\psfrag{d}{\tiny  $y_2$}
\psfrag{e}{ \small $a$}
\includegraphics[scale = 0.65]{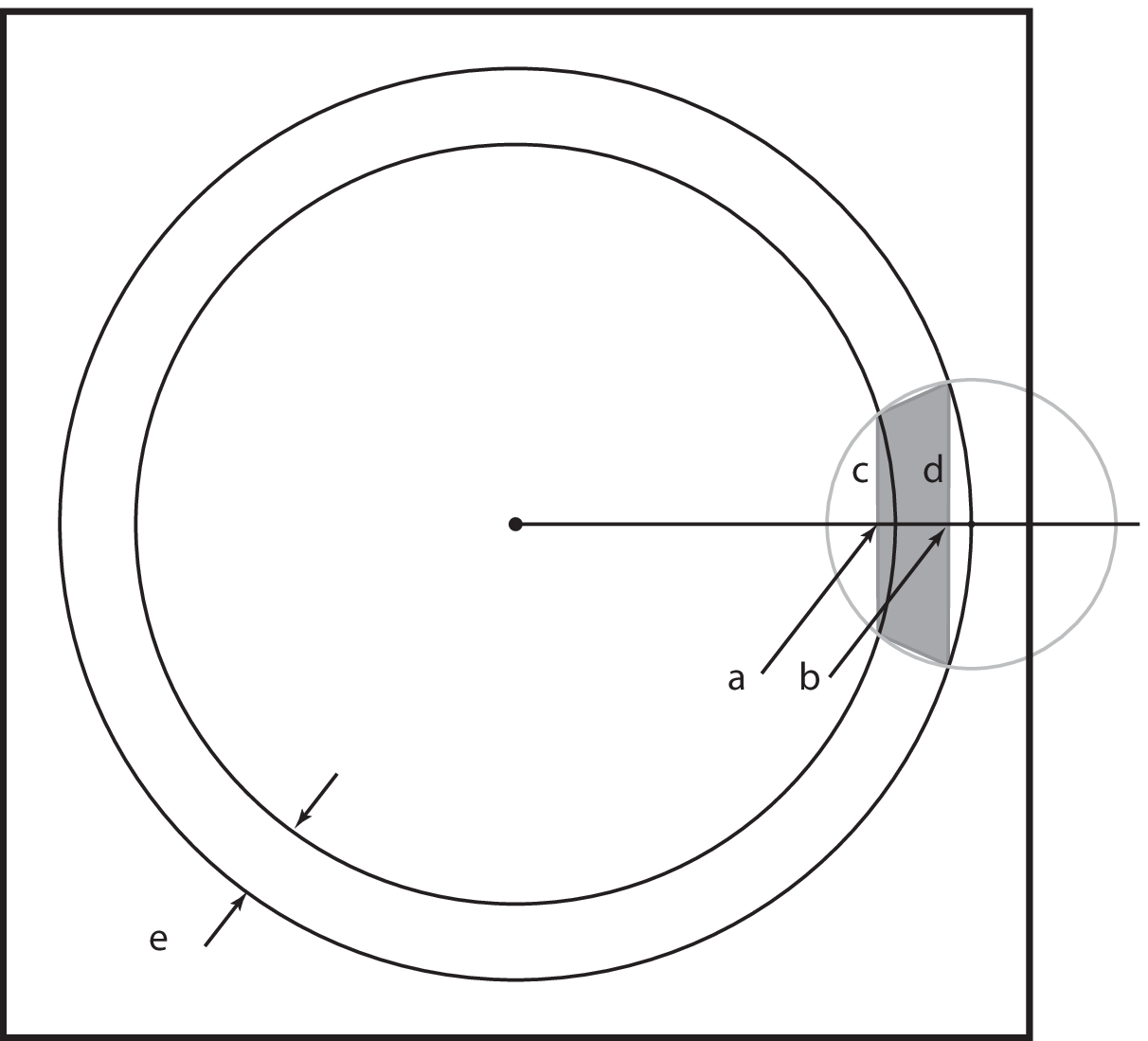}
\label{fig:lower_bound2}}
\caption{Evaluation of lower bound for $A(r)$. In \subref{fig:lower_bound1} we assume that $\delta_n \leq a$ whereas in \subref{fig:lower_bound2} we take $\delta_n > a$. In both cases the minimum intersection is achieved when the center of $\delta_n$ circle is on the exterior boundary of the region.}
\end{figure}

\textbf{Case 1 ($\delta_n \leq a$):}\\
In this case the minimum area of the intersection is achieved when the center of the circle is on the exterior boundary of the region as shown in Fig.~\ref{fig:lower_bound1}. 
In this case, one can show that, 
\begin{equation}
\label{eq:lower_bound1}
A(r) \geq \frac{\pi\delta_n^2}{4}\,.
\end{equation}

\textbf{Case 2 ($\delta_n > a$): }\\
In this case, wherever the center of the circle is, it will have intersection with both bounding circles. Thus, the minimum area is achieved when the center of the circle is on the exterior boundary as in Fig.~\ref{fig:lower_bound2}, where
\begin{equation*}
\begin{matrix}
\begin{aligned}
x_1 = \frac{r_2^2 - \delta_n^2+r_1^2}{2r_2}, \quad & y_1= \frac{1}{r_2}\sqrt{(\delta_n^2 - a^2)(4r_0^2 - \delta_n^2)}\\
x_2 = \frac{r_2^2 - \delta_n^2+r_2^2}{2r_2}, \quad & y_2= \frac{1}{r_2}\sqrt{\delta_n^2(4r_2^2 - \delta_n^2)}\,.
\end{aligned}
\end{matrix}
\end{equation*}

Thus, we will have
%\begin{equation*}
\begin{align}
A(r) &\geq \frac{y_1+y_2}{2}(x_2-x_1)\nonumber\\
&= \frac{\sqrt{(\delta_n^2 - a^2)(4r_0^2 - \delta_n^2)} +\sqrt{\delta_n^2(4r_2^2 - \delta_n^2)} }{2r_2}\cdot \frac{r_2^2 - r_1^2}{2r_2}\nonumber\\
&\geq \frac{\sqrt{\delta_n^2(4r_2 - \delta_n^2)}}{2r_2}\cdot \frac{r_2^2 - r_1^2}{2r_2}= \delta_n\sqrt{(r_2 - \frac{1}{4}\delta_n^2)}\frac{r_2^2 - r_1^2}{2r_2^2}\nonumber
\end{align}
%\end{equation*}
If we assume that $r_2 \geq \frac{1}{\sqrt{2}}\delta_n$, which is a reasonable assumption according to the problem statement, we will have
\begin{align}
A(r) &{\geq} \frac{1}{2}\delta_n^2 \frac{r_2^2 - r_1^2}{2r_2^2}\geq \frac{a\,r_0}{2(r_0+a)^2} \delta_n^2\,.\label{eq:lower_bound2}
\end{align}

 Combining \eqref{eq:lower_bound1} and \eqref{eq:lower_bound2}, we can find the lower bound for $A(r)$ as
\begin{equation*}
\begin{aligned}
A(r) &\geq \min(\frac{\pi}{4},\frac{a\,r_0}{2(r_0+a)^2}) \delta_n^2 = \frac{a\,r_0}{2(r_0+a)^2}\delta_n^2\,.
\end{aligned}
\end{equation*}
Thus, 
\begin{equation*}
\begin{aligned}
q &= \int_{r_1}^{r_2}\frac{2\pi r}{\pi(r_2^2 - r_1^2)}p_2(r)dr=   \int_{r_1}^{r_2}\frac{2\pi r}{\pi(r_2^2 - r_1^2)}\frac{A(r)}{\pi(r_2^2 - r_1^2)}dr\geq \frac{\delta_n^2}{4\pi(r_0+a)^2}\,.
\end{aligned}
\end{equation*}
From the above calculations, we have that $  \frac{\delta_n^2}{4\pi(r_0+a)^2} p_n\,n\leq \E[Y_i]\leq \frac{1}{r_0}\delta_np_n\, n$. Applying the Chernoff bound to $Y_i$, we have
\begin{equation*}
\prob\Big( Y_i > (1+\alpha)\E[Y_i]\Big) \leq 2^{-(1+\alpha)\E[Y_i]}\,.
\end{equation*}
In other words
\begin{equation*}
\prob\Big( Y_i > (1+\alpha)\frac{1}{r_0}\delta_np_n\,n\Big) \leq 2^{-(1+\alpha) \frac{\delta_n^2}{4\pi(r_0+a)^2} p_n\,n}\,.
\end{equation*}
Applying the union bound, we get
\begin{equation*}
\begin{aligned}
\prob\Big(\max_{i \in [n]} Y_i > (1+\alpha)\frac{1}{r_0}\delta_np_n n\Big) &\leq n 2^{-(1+\alpha) \frac{\delta_n^2}{4\pi(r_0+a)^2} p_n\,n}\leq 2^{-\left((1+\alpha) \frac{\delta_n^2}{4\pi(r_0+a)^2} p_n\,n - \log_2n\right)}\,.
\end{aligned}
\end{equation*}
By the assumption that $\delta_np_n = \Omega(r_0 \sqrt{\log_2n/n})$, there exists constants $c$ and $N$, such that $\delta_n^2p_n \geq cr_0^2\log_2n/n$, for $n\geq N$. Define a positive parameter $\beta$ such that $1+\beta = \frac{c(1+\alpha)r_0^2}{4\pi (r_0+a)^2}$. Then we will have
\begin{equation*}
\prob\Big(\max_{i\in[n]}Y_i > \frac{4\pi(1+\beta)}{cr_0^3}(r_0+a)^2\delta_np_n\,n\Big) \leq n^{-\beta}\,.
\end{equation*}
Finally with probability $1-n^{-\beta}$, 
\begin{equation*}
\begin{aligned}
\|\cP_E(\bD^s)\|_2 &\leq \frac{4\pi(1+\beta)}{c}\delta^3\left(\sqrt{\frac{\log n}{n}}\right)^3p\,n = C(r_0+a)^2\delta^3\left(\sqrt{\frac{\log n}{n}}\right)^3p\,n\,.
\end{aligned}
\end{equation*}
This finishes the proof of Lemma \ref{lem:2norm}.\myqed
\end{document}